\documentclass[10pt,twocolumn,letterpaper]{article}

\usepackage[pagenumbers]{cvpr} 

\usepackage{graphicx}
\usepackage{amsmath}
\usepackage{amssymb}
\usepackage{booktabs}
\usepackage{multirow}

\usepackage[pagebackref,breaklinks,colorlinks]{hyperref}

\usepackage[capitalize]{cleveref}
\crefname{section}{Sec.}{Secs.}
\Crefname{section}{Section}{Sections}
\Crefname{table}{Table}{Tables}
\crefname{table}{Tab.}{Tabs.}

\def\cvprPaperID{4794} 
\def\confName{CVPR}
\def\confYear{2024}

\begin{document}

\title{Semantic-Preserved Point-based Human Avatar}

\author{Lixiang Lin \quad Jianke Zhu \\
ZheJiang University\\
{\tt\small \{lxlin,jkzhu\}@zju.edu.cn}
}
\maketitle

\begin{abstract}

To enable realistic experience in AR/VR and digital entertainment, we present the first point-based human avatar model that embodies the entirety expressive range of digital humans. We employ two MLPs to model pose-dependent deformation and linear skinning (LBS) weights. The representation of appearance relies on a decoder and the features that attached to each point. In contrast to alternative implicit approaches, the oriented points representation not only provides a more intuitive way to model human avatar animation but also significantly reduces both training and inference time. Moreover, we propose a novel method to transfer semantic information from the SMPL-X model to the points, which enables to better understand human body movements. By leveraging the semantic information of points, we can facilitate virtual try-on and human avatar composition through exchanging the points of same category across different subjects. Experimental results demonstrate the efficacy of our presented method. Code will be made publicly available at \href{https://github.com/l1346792580123/spa}{https://github.com/l1346792580123/spa}

\end{abstract}

\section{Introduction}
\label{sec:intro}

Nonverbal human elements such like body posture, facial expressions, and appearance are essential to human communication. The presence of these nonverbal cues highlights the need of capturing the complete essence of human expressiveness so that we can achieve immersive and lifelike experiences in AR/VR. Currently, the available solutions can be broadly categorized into two groups, including model-based methods~\cite{DBLP:journals/tog/SMPL15, DBLP:conf/cvpr/SMPLX19, DBLP:conf/eccv/smplify16, DBLP:conf/cvpr/HMR18} and model-free approaches~\cite{DBLP:conf/iccv/PIFu19, DBLP:conf/cvpr/pifuhd20, DBLP:conf/iccv/snarf21, DBLP:conf/cvpr/scanimate21, shen2023xavatar}. Unfortunately, both of them have their own limitations and drawbacks.

\begin{figure}
	\centering
    \includegraphics[width=1.0\linewidth]{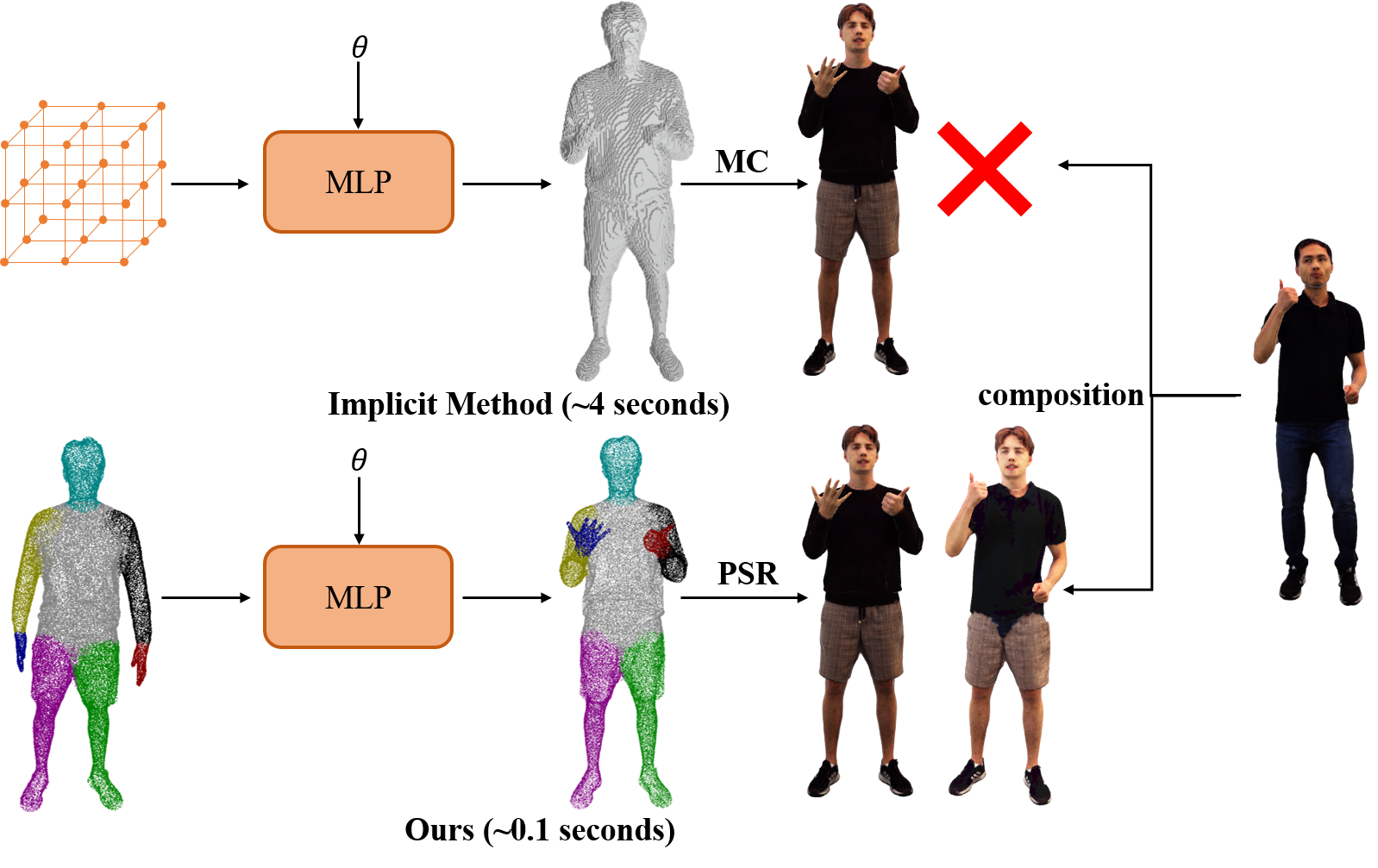}
	\caption{The conventional implicit approach involves traversing each position in the space to obtain the SDF or occupancy value, and then passing through the marching cube (MC)~\cite{DBLP:conf/siggraph/Marchingcubes87} to get the final mesh. While our method intuitively animate the oriented points. Our points representation can produce the final mesh through Poisson reconstruction (PSR)~\cite{DBLP:conf/sgp/psr06} in no time while preserving the semantic information of the human body. Points with semantics allow us to exchange points of different individuals to achieve human avatar composition.}
	\label{fig:teaser}
\end{figure}

The development of parametric body models, such like SMPL family~\cite{DBLP:journals/tog/SMPL15, DBLP:conf/cvpr/SMPLX19}, has played a crucial role in digital human modeling. These models utilize PCA shape parameters and Euler angle of joints to represent human in variety poses. While the fixed topology and limited resolution of the 3D mesh restrict their flexibility and prevent modeling of clothing or hair. Consequently, capturing the complete appearance of humans becomes challenging with these models, especially when considering clothed bodies.

Implicit representations offer a promising solution to overcome these limitations. Chen \textit{et al}~\cite{DBLP:conf/iccv/snarf21} combined the continuous implicit functions with learned forward skinning to build articulate human avatars, which can produce reasonable results for arbitrary poses. Saito \textit{et al}~\cite{DBLP:conf/cvpr/scanimate21} proposed an end-to-end framework that aims to learn geometric cycle-consistency while considering both posed and unposed shapes in a weakly supervised manner. Shen \textit{et al}~\cite{shen2023xavatar} incorporated part-aware initialization strategy on the top of SNARF to reduce the ambiguity in iterative root finding. Moreover, hand pose, facial expressions and appearance are included to model expressive human avatar in a holistic fashion. Although having achieved promising results, implicit method still suffers some problems. The implicit representation is not straightforward, as it requires forward inference to derive geometric information, such as occupancy, SDF, etc. It takes long time to obtain final mesh since all grid points in space are passed to MLP to query geometric information. Moreover, implicit representations cannot maintain semantic information. To depict the same individual in various outfits, it requires to train distinct models for each attire variation.

To tackle these problems, we propose an oriented points-based method to model human avatar. As depicted in Fig.~\ref{fig:teaser}, our oriented points representation proficiently models human avatars with minimal inference time. We introduce a novel approach to transfer the semantic information from SMPL-X model~\cite{DBLP:conf/cvpr/SMPLX19} to our oriented points representation. Enriching points with semantics enhances our ability to analyze the meaning of human movements. Additionally, we combine points from various subjects based on their semantic attributes to generate new human avatars, a valuable application for virtual try-on scenarios. Oriented points representation offers a more intuitive alternative comparing to implicit methods. By leveraging the Poisson Surface Reconstruction technique~\cite{DBLP:conf/sgp/psr06}, we can generate the final mesh instantly, whereas implicit methods may require several seconds. We integrate two MLPs to capture pose-dependent deformations and linear blend skinning~(LBS) weights. These oriented points are then skinned using SMPL-X pose parameters, enabling us to represent a wide range of human poses. We affix a trainable neural texture to each point as appearance representation. This neural texture is decoded through an MLP to determine the final color. The texture of the vertices on the reconstructed mesh is derived as a weighted average of their k-nearest points.

In summary, the main contributions of our paper are in the following.
\begin{itemize}
    \item We present the first point based human avatar that can capture body pose, hand pose, facial expressions and appearance.
    \item We propose a novel method to transfer the semantic information of SMPL-X model to the point-based avatar, which enables some real-word application such like virtual try-on.
    \item Experimental evaluations against state-of-the-art methods demonstrate the accuracy and efficiency of our proposed method.
\end{itemize}


\begin{figure*}
    \centering
    \includegraphics[width=1.0\linewidth]{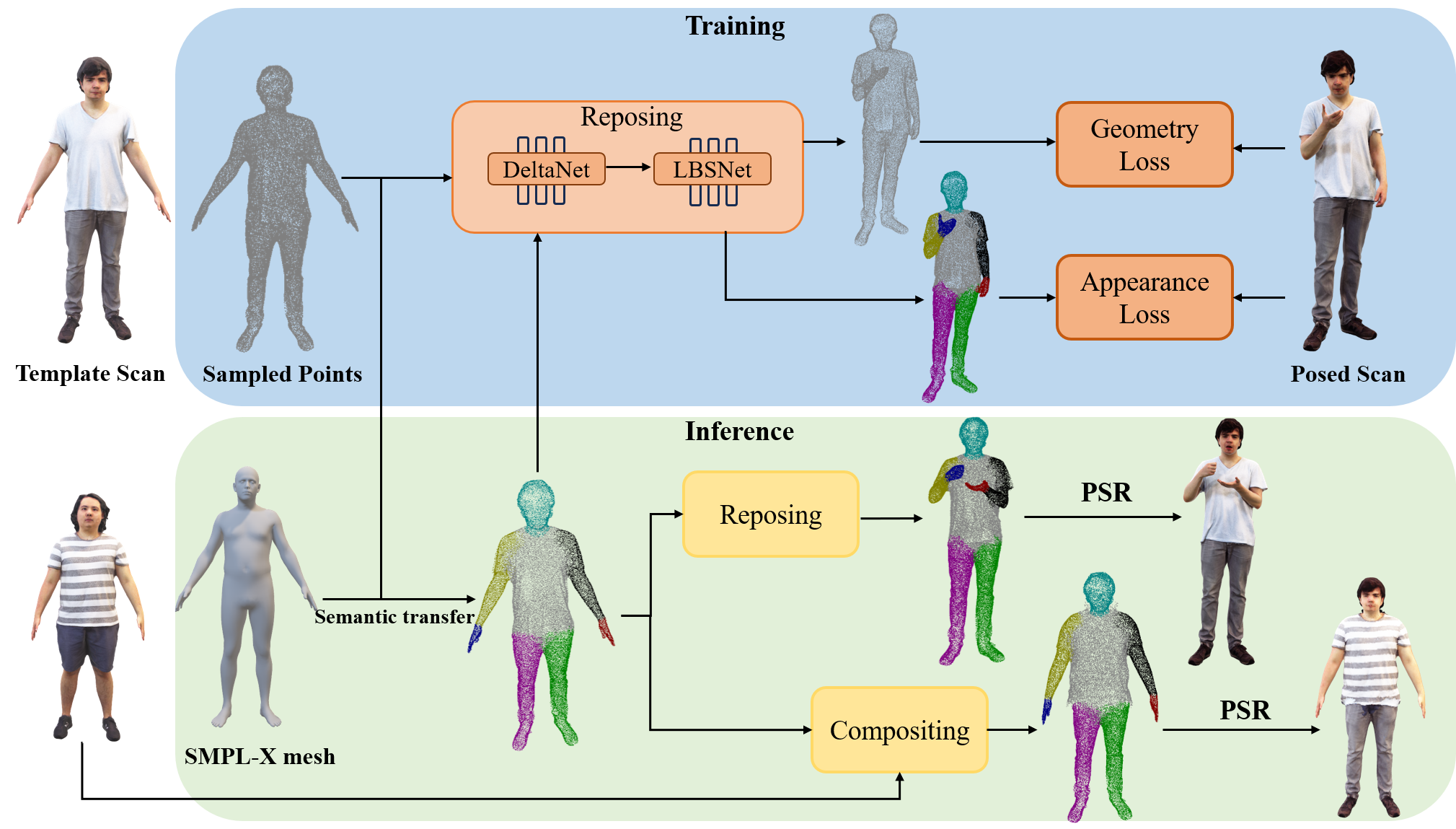}
    \caption{\textbf{Method Overview}. We select a scan from training set as template scan. During training, the sampled points are transformed to pose space through DeltaNet and LBSNet, and then the loss is computed by comparing it with points sampled from the posed scan. During inference, we transfer the semantic information of SMPL-X mesh to sampled points. The sampled points can be reposed according to SMPL-X parameters, and can also be composited with other points to generate a new avatar. For clearness, we omit the Poisson reconstruction from points to mesh.}
    \label{fig:overview}
\end{figure*}

\section{Related Works}

\subsection{Parametric Human Models}
Parametric human models~\cite{DBLP:journals/tog/SCAPE05,DBLP:journals/tog/SMPL15,DBLP:conf/cvpr/TotalCapture18,DBLP:conf/cvpr/SMPLX19} try to represent the shape and pose of animatable people with a few parameters. Some models focus on a part of human like hands~\cite{DBLP:journals/tog/mano17} or faces~\cite{DBLP:conf/siggraph/BlanzV99,DBLP:journals/tog/LiBBL017}. Parametric models are favored due to their seamless integration with existing computer graphics pipelines and the advantage of a compact parameter space that facilitates effective learning. Traditional approaches~\cite{DBLP:conf/eccv/smplify16,DBLP:conf/cvpr/MonocularTotalCapture19,DBLP:conf/cvpr/SMPLX19} rely on nonlinear optimization solver to deduce the reasonable parameters from 2D keypoints, which are usually computationally inefficient. Many methods~\cite{DBLP:conf/cvpr/HMR18,DBLP:conf/iccv/SPIN19,DBLP:conf/eccv/ChoutasPBTB20,DBLP:conf/3dim/FengCBTB21,DBLP:conf/iccv/KocabasHHB21,DBLP:conf/iccv/ZhangTZOL0S21} adopt deep neural network to directly predict parameters from a single RGB image. Furthermore, the widespread adoption of graph convolutional networks~\cite{DBLP:conf/eccv/Pose2Mesh20, DBLP:conf/eccv/COMA18} has made it possible to efficiently reconstruct human meshes from 2D keypoints. Because of the constrained expressive capacity of parametric human models, these methods typically generate a representation of the human body with minimal clothing, neglecting the finer details of garments, hair, and other accessories. Some efforts have been devoted to representing clothing as an offset layer from the underlying body~\cite{DBLP:conf/3dim/DetailedHumanAvatars18,DBLP:conf/cvpr/PeopleinClothing19,DBLP:conf/cvpr/HMD19,DBLP:journals/tog/ClothCap17}. While the introduction of an offset layer has enhanced their representation capabilities, the gap between these SMPL+D methods and real-world human body is difficult to bridge.

\subsection{Implicit Human Models}
Implicit human models utilize voxels or implicit functions to describe geometry, circumventing the need for explicit models and offering a much larger solution space for capturing intricate details. Voxels representations~\cite{DBLP:conf/eccv/BodyNet18, DBLP:conf/iccv/DeepHuman19} demand cubic memory, which hinders these methods from obtaining the high resolution reconstruction results. Rather than relying on voxels, implicit functions parameterized by MLPs present a memory efficient way for representing geometry. Given a spatial point, signed distance fields (SDF)~\cite{DBLP:conf/cvpr/deepsdf19}, occupancy~\cite{DBLP:conf/cvpr/occnet19,DBLP:conf/iccv/PIFu19,DBLP:conf/cvpr/pifuhd20}, or density~\cite{DBLP:conf/eccv/nerf20} can be estimated by MLPs, and triangulate mesh can be extracted by marching cube algorithm~\cite{DBLP:conf/siggraph/Marchingcubes87}. Although implicit functions approaches offer memory efficient geometry representation, more training and inference time are required as a trade-off. Training implicit function-based methods can be time-consuming, as the entire MLP is updated during each iteration.  In the inference phase, high-resolution grid points are fed into the implicit functions to retrieve geometric information. A recent work~\cite{Peng2021SAP} called SAP introduces a differentiable Poisson solver to bridge oriented point clouds, implicit indicator functions, and meshes altogether. SAP employs light-weight oriented point clouds to represent shapes, which greatly reduces training time and inference time. 

\subsection{Human Avatar}

Numerous endeavors have been undertaken to create an animatable human avatar by amalgamating parametric human models and implicit human models~\cite{DBLP:conf/eccv/nasa20, DBLP:conf/iccv/snarf21, DBLP:conf/cvpr/scanimate21, shen2023xavatar}. NASA~\cite{DBLP:conf/eccv/nasa20} exploits per body-part occupancy networks~\cite{DBLP:conf/cvpr/occnet19} to represent articulated human bodies, which can potentially introduce artifacts for unseen poses. SNARF~\cite{DBLP:conf/iccv/snarf21} proposes a forward warping field and incorporates SMPL~\cite{DBLP:journals/tog/SMPL15} skeleton to learn pose-independent skinning. SCANimate~\cite{DBLP:conf/cvpr/scanimate21} utilizes a weakly supervised learning method that takes raw 3D scans and turns them into an animatable avatar without surface registration. X-Avatar~\cite{shen2023xavatar} proposes a part-aware initialization and sampling strategies to improve the correspondence search in iterative root finding. SMPL-X~\cite{DBLP:conf/cvpr/SMPLX19} skeleton is adopted to model body pose, hand pose, facial expressions together. Despite achieving promising results, these methods still face several challenges. The process of acquiring the final mesh is time-consuming, as it involves passing all grid points in space through the MLP to retrieve geometric information. Additionally, implicit representations struggle to preserve semantic information. Also, there are some works that model dynamic humans from multiview videos~\cite{peng2021neural, DBLP:conf/iccv/animatenerf21}. These methods utilize image pixels as their optimization target, which can effectively generate images of humans in various poses. Nevertheless, they struggle to capture accurate geometry and exhibit weak generalization when faced with unseen poses due to the limitations of volume rendering and density representation.


\section{Methods}

In this section, we introduce our oriented point-based method for human avatars modeling, which preserves semantic information. Fig.~\ref{fig:overview} shows the overview of our proposed method. We firstly introduce the oriented points shape representation in Section~\ref{sec:shape_representation}. Then, we delve the representations of pose-dependent deformation and skinning weights in Section~\ref{sec:lbs_deform}. Moreover, we elaborate on how we transfer semantic information from SMPL-X model to oriented points in Section~\ref{sec:semantic_transfer}. Additionally, we explore the appearance representation in Section~\ref{sec:appearance}. The specific details of our implementation can be found in Section~\ref{sec:implement}.

\subsection{Shape Representation}\label{sec:shape_representation}

Differently from those implicit methods~\cite{DBLP:conf/iccv/snarf21,DBLP:conf/cvpr/scanimate21,shen2023xavatar}, we employ oriented point clouds $S = \{ \mathbf{x} \in \mathbb{R}^3, \mathbf{n} \in \mathbb{R}^3 \}$ as shape representation. Oriented point clouds representation is interpretable, lightweight and fast comparing to implicit representation. Implicit methods store the mesh within the MLP and necessitate forward inference to query geometric information at individual positions. We represent the mesh explicitly through the positions and normals of points. In implicit methods, it is time consuming to query all grid positions for each pose in order to acquire occupancy or SDF value and subsequently employ marching cube~\cite{DBLP:conf/siggraph/Marchingcubes87} to generate final mesh. In contrast, by taking advantage of oriented point clouds representation, we can intuitively observe pose-dependent deformation and obtain final mesh through Poisson Surface Reconstruction~\cite{DBLP:conf/sgp/psr06} very efficiently.

\subsection{Oriented Points Deformation}\label{sec:lbs_deform}

\begin{figure}
    \centering
    \includegraphics[width=1.0\linewidth]{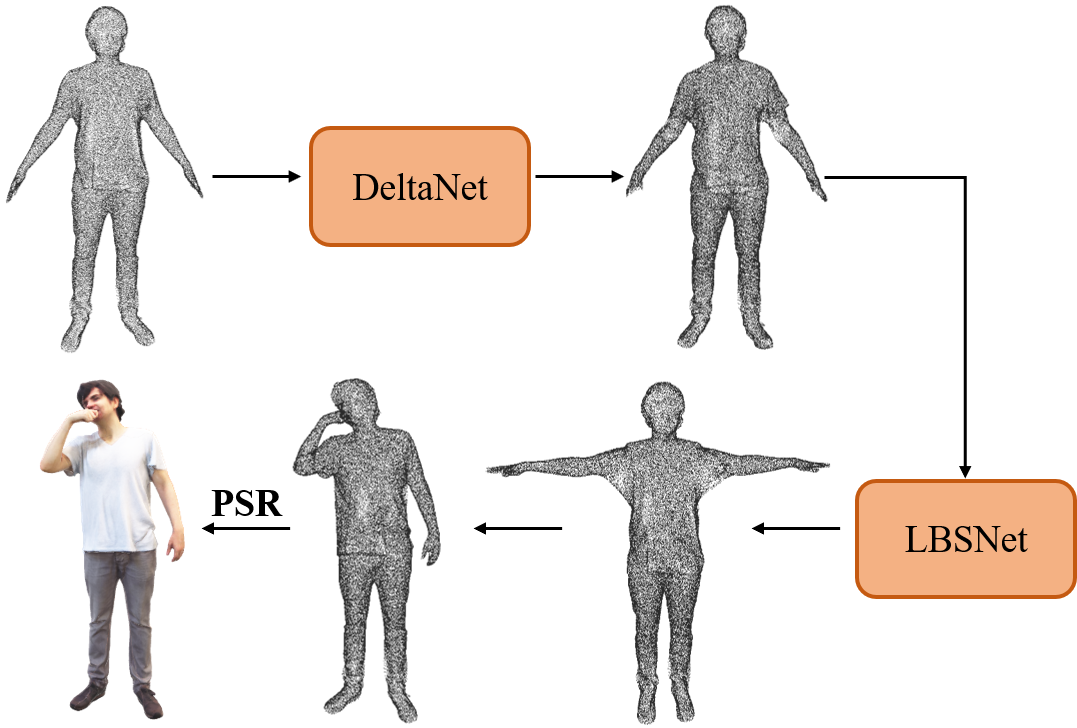}
    \caption{\textbf{Points Deformation}. The sampled points are employed pose-dependent deformation through DeltaNet. Then we estimate LBS weights through LBSNet. Points are firstly skinned to canonical space and then to pose space. Mesh can be extracted through Poisson reconstruction.}
    \label{fig:deform}
\end{figure}

As shown in Fig.~\ref{fig:deform}, the oriented points undergo an initial deformation step using DeltaNet, followed by the estimation of skinning weights through LBSNet.  Subsequently, the points are linear skinned to the pose space according to the skeleton of SMPL-X model. We start by transforming the points from the template space to the canonical space and subsequently into the pose space.

Similar to ~\cite{DBLP:conf/cvpr/pointavatar23}, we utilize a coordinate-based MLP to map each sampled point $\mathbf{x}$ to both an offset and an Euler angle, enabling the modeling of pose-dependent deformation.
\begin{equation}
    f_d : \mathbb{R}^3 \times \mathbb{R}^{|\theta_b| + |\psi|} \rightarrow \Delta \in \mathbb{R}^3, \Theta \in \mathbb{R}^3
\end{equation}
\begin{equation}
    \mathbf{x}_c = \mathbf{x} + \Delta
\end{equation}
\begin{equation}
    \mathbf{n}_c = R(\Theta) \mathbf{n}
\end{equation}
where the offset $\Delta$ and the Euler angle $\Theta$ are conditioned by body pose $\theta_b$, facial expression $\psi$ and points positions $\mathbf{x}$. $R(\Theta)$ represents the rotation matrix computed from Euler angle. We incorporate position encoding~\cite{DBLP:conf/eccv/nerf20} to model high-frequency details. The sampled points and normals are firstly displaced by the pose-dependent offset and rotation matrix, which are further deformed to pose space according to linear blend skinning.

Similar to previous work~\cite{DBLP:conf/iccv/snarf21,DBLP:conf/cvpr/imavatar22, shen2023xavatar}, we employ another MLP to represent the skinning weight field in the template space
\begin{equation}
    f_{s} : \mathbb{R}^3  \rightarrow \mathbb{R}^{N_b},
\end{equation}
where $N_b$ represents the number of bones, including body, finger and face bones. We incorporate linear blend skinning to model the pose transformation. Specifically, each bone $\text{b}_i$ corresponds to a weight $w_i$ and a bone transformations $\text{B}_i$. $\text{B}_i$ is computed by pose $\theta$ and kinematic tree of SMPL-X model. For each point $\mathbf{x}_c$ in template space, the corresponding point $\mathbf{x}_d$ in pose space can be determined by
\begin{equation}
    \begin{aligned}
    f_s(\mathbf{x}_c) = \{ w_1(\mathbf{x}_c),...,w_{N_b}(\mathbf{x}_c) \} \\ 
    \text{w.r.t.} \quad w_i \geq 0 \ \text{and} \  \sum\nolimits_i w_i = 1
    \end{aligned}
\end{equation}
\begin{equation}
    \mathbf{x}_d = \sum_{i=1}^{N_b} w_i(\mathbf{x}_c) \text{B}_i {\text{B}^c_i}^{-1} \mathbf{x}_c
\end{equation}
\begin{equation}
    \mathbf{n}_d = \sum_{i=1}^{N_b} w_i(\mathbf{x}_c) \text{B}_i {\text{B}^c_i}^{-1} \mathbf{n}_c
\end{equation}
where $\text{B}_i$ and $\text{B}^c_i$ represent the bone transformations corresponding to pose $\theta$ and template pose $\theta_t$, respectively. Since the linear skinning of SMPL-X model is performed in canonical space, we first transform points to canonical space and then to pose space. The deformed points $\mathbf{x}_d$ are employed to calculate loss with the points sampled from pose scan $\mathbf{x}_p$
\begin{equation}
    \mathcal{L} = \mathcal{L}_{chamfer} + \mathcal{L}_{EMD} + \mathcal{L}_{normal} + \mathcal{L}_{reg}.
\end{equation}
We employ regular Chamfer distance as preliminary loss function as follows
\begin{equation}
    \begin{split}
        \mathcal{L}_{chamfer} & = \frac{1}{|\mathbf{x}_d|} \sum\limits_{i \in |\mathbf{x}_d|} \min\limits_{j \in |\mathbf{x}_p|} (||\mathbf{x}_d^i - \mathbf{x}_p^j||_2^2) \\
        & + \frac{1}{|\mathbf{x}_p|} \sum\limits_{j \in |\mathbf{x}_p|} \min\limits_{i \in \mathbf{x}_d} (||\mathbf{x}_p^j - \mathbf{x}_d^i||_2^2).
    \end{split}
\end{equation}
However, only chamfer distance may cause uneven density distribution and blurred details, which affects the quality of the mesh generated by Poisson reconstruction. We additionally employ Earth Mover's Distance (EMD)~\cite{DBLP:conf/aaai/msn020} to avoid this situation
\begin{equation}
    \mathcal{L}_{EMD} = \min\limits_{\phi: \mathbf{x}_d \rightarrow \mathbf{x}_p} \frac{1}{|\mathbf{x}_d|} \sum\limits_{i \in \mathbf{x}_d} ||x_d^i - \phi(x_d^i)||,
\end{equation}
where $\phi$ is a bijection. EMD loss minimizes the dissimilarity between two point sets and force $\mathbf{x}_d$ to be evenly distributed since $\mathbf{x}_p$ are sampled evenly. In order to capture clothing details, normal consistency loss is employed
\begin{equation}
    \mathcal{L}_{normal} = \frac{1}{|\mathbf{x}_d|} \sum\limits_{i \in \mathbf{x}_d} 1 - \cos(\mathbf{n}_d^i, \phi(\mathbf{n}_d^i))
\end{equation}
where $n_d^i$ is the normal vector of point $\mathbf{x}_d^i$. Cosine similarity enforces normal consistency. Moreover, we regularize the LBS weights to be close to those of SMPL-X and constrain the offset to tend to zero.
\begin{equation}
    \begin{split}
        \mathcal{L}_{reg} &= \frac{1}{|\mathbf{x}_{x|}} \sum\limits_{i=1}^{N_b} || {w_i(\mathbf{x}_{x})} - w_i(\mathbf{x}_{x})^* ||_2^2 \\
        &+ \frac{1}{|\mathbf{x}|}||f_d(\mathbf{x};\theta_b,\psi)||_2^2
    \end{split}
\end{equation}
where $\mathbf{x}_x$ represents the vertices of registered SMPL-X mesh in template space. $w_i(\mathbf{x}_{x})^*$ is the LBS weights defined by SMPL-X model.


\begin{figure}
    \centering
    \includegraphics[width=1.0\linewidth]{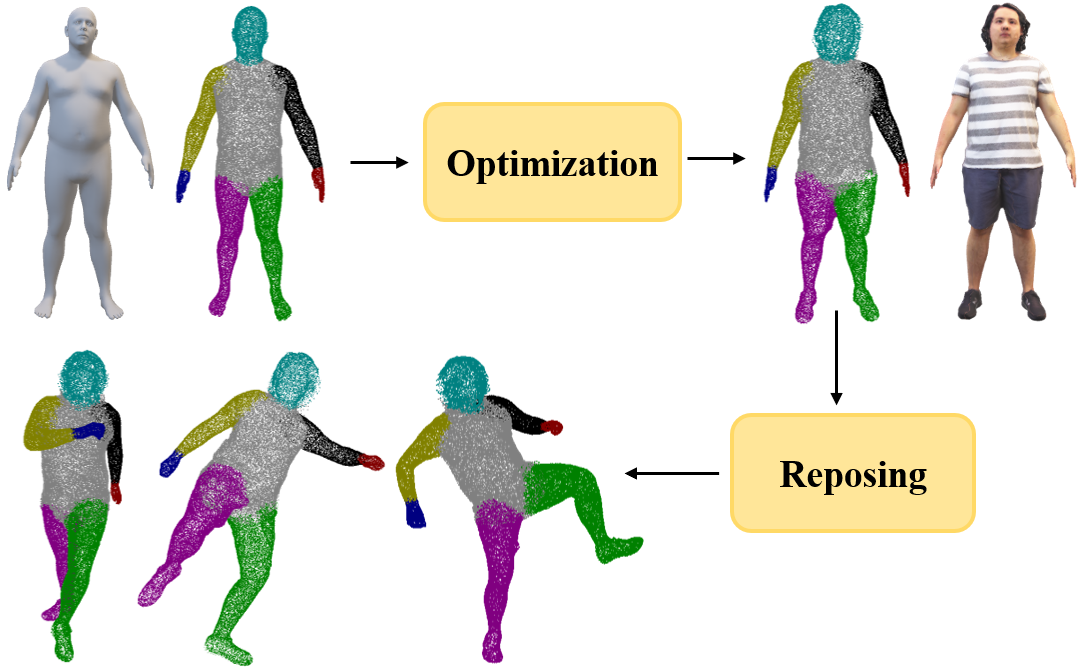}
    \caption{\textbf{Semantic Transfer}. We divide SMPL-X models into 8 parts. We optimize the point cloud with semantic information sampled from SMPL-X mesh according to template scan. Then the sampled point cloud can be reposed like SMPL-X model. }
    \label{fig:transfer}
\end{figure}

\subsection{Semantic Transfer}\label{sec:semantic_transfer}
As shown in Fig.~\ref{fig:transfer}, once the geometry training is done, we transfer the semantic information of SMPL-X mesh to point clouds. SMPL-X model~\cite{DBLP:conf/cvpr/SMPLX19} is a parametric human model that is parameterized by shape $\beta$, pose $\theta$ and facial expressions $\psi$. 

The semantic information of SMPL-X model is predefined, since the topology of SMPL-X is fixed. We categorize the SMPL-X points into 8 parts: \{head, body, left arm, right arm, left hand, right hand, left leg, right leg\}. The label of SMPL-X faces are derived from the corresponding points. We fix the relative position of sampled points to the faces so that the sampled points can maintain correspondence and semantic information. To align the template scan, sampled point clouds $\mathbf{x}_s$ are employed in the same Chamfer loss and EMD loss described in Section~\ref{sec:lbs_deform}. After transferring the points, the semantic points can be deformed according to different pose parameters. By leveraging the correspondences, we are able to swap point clouds belonging to the same label to composite avatar. Once the semantic information of each point is determined, we freeze the DeltaNet and LBSNet and optimize the appearance feature.

\subsection{Appearance Representation}\label{sec:appearance}

Once the geometry training and semantic transfer are completed, we attach a learnable feature to each point, serving as its appearance representation. We first train an autoencoder for color representation as follows
\begin{equation}
    c = D(E(c)).
\end{equation}
Subsequently, we freeze the decoder and optimize the feature. We transform the semantic points to pose space and perform loss calculation with sampled points at point level. For each sampled point $\mathbf{x}_p^i$, we aggregate features from k-nearest semantic points. We employ inverse distance weighting (IDW) to weight features
\begin{equation}
    \mathcal{F}_p^i =  \frac{ \sum_{j \in \mathcal{N}(i)} d_i^j \mathcal{F}_s^j}{\sum_{j \in \mathcal{N}(i)} d_i^j },
\end{equation}
where $\mathcal{F}_s^j$ indicates the learnable feature of point $\mathbf{x}_p^j$. $\mathcal{N}(i)$ represents the k-nearest points of $\mathbf{x}_p^i$. $d_i^j$ is the distance between the sampled point $\mathbf{x}_p^i$ and semantic point $\mathbf{x}_s^j$. The aggregated feature is then passed to the decoder $D$ to obtain the final color and compute the loss as below
\begin{equation}
    \mathcal{L}_{color} = \frac{1}{|\mathbf{x}_p|} \sum\limits_{i \in |\mathbf{x}_p|}||D(\mathcal{F}_p^i) - c_p^i||_2^2,
\end{equation}
where $c_p^j$ is the color of point $\mathbf{x}_p^j$.

\subsection{Implementation Details}\label{sec:implement}

We implement the proposed method in PyTorch~\cite{NEURIPS2019_pytorch}. We train the points deformation with 100 epochs and train the appearance feature for another 20 epochs. At each iteration, we sample $51,200$ points from the template scan and pose scan, respectively. We employ Adam optimizer~\cite{DBLP:journals/corr/adam14} for optimization. The learning rate for DeltaNet, LBSNet, and feature are $5 \times 10^{-4}$, $1 \times 10^{-4}$, $1 \times 10^{-3}$, respectively. Pose $\theta_b$ and facial expressions $\psi$ are firstly encoded to $16$ dimensional features through network, and then concatenated with points‘ positions as input. DeltaNet consists of 8 layers with hidden layers of width 512, and a skip connection from the input to the middle layer. LBSNet consists of 5 layers, with hidden layers of width 128. We also employ position encoding in LBSNet to capture complex hand and face poses. The dimension of position encoding used by DeltaNet and LBSNet is 4. We incorporate hierarchical softmax introduced in SNARF~\cite{DBLP:conf/iccv/snarf21} to better estimate LBS weights. The weight of soft blend is $20$. Color autoencoder consists of 8 layers with hidden layers 256. The dimension of feature is 16. We employ the Chamfer loss and EMD loss from 
Kaolin~\cite{KaolinLibrary} and~\cite{DBLP:conf/aaai/msn020}, respectively. We set the weights of the losses to $\lambda_{chamfer}=5000, \lambda_{EMD}=5000, \lambda_{normal}=1, \lambda_{reg}=100, \lambda_{color}=10$, respectively.

\section{Experiments}

In this section, we present the experimental results on GRAB~\cite{DBLP:conf/eccv/grab20} and X-Humans~\cite{shen2023xavatar}. We compare our proposed method with the state-of-the-art implicit methods. Moreover, we conduct ablation studies to evaluate the effectiveness of proposed losses. Furthermore, we demonstrate the effectiveness of point representation with semantics in avatar composition. 

\begin{table}
    \centering
    \caption{Quantitative results on GRAB dataset}
    \resizebox{1.00\linewidth}{!}{
    \begin{tabular}{lcc|cc|cc|cc}
    \toprule
    \multirow{2}{*}{Method} & \multicolumn{2}{c}{CD$\downarrow$}  & \multicolumn{2}{c}{CD-MAX $\downarrow$}  & \multicolumn{2}{c}{NC $\uparrow$} & \multicolumn{2}{c}{IoU $\uparrow$ } \\
    \cline{2-9} & All & Hands & All & Hands & All & Hands & All & Hands\\
    \midrule
    SCANimate~\cite{DBLP:conf/cvpr/scanimate21} & 2.60 & 8.39 & 54.75 & 54.22 & 0.967 & 0.760 & 0.941 & 0.569 \\
    SNARF~\cite{DBLP:conf/iccv/snarf21} & 1.37 & 5.13 & 33.86 & 33.51 & 0.977 & 0.818 & 0.967 & 0.739 \\
    X-Avatar~\cite{shen2023xavatar} & 0.94 & 0.79 & 21.43 & 4.79 & \textbf{0.985} & 0.957 & 0.991 & 0.895 \\
    Ours & \textbf{0.85} & \textbf{0.77} & \textbf{19.76} & \textbf{4.71} & \textbf{0.985} & \textbf{0.962} & \textbf{0.993} & \textbf{0.901} \\
    \bottomrule
    \end{tabular}}
    \label{tab:grab}
\end{table}


\begin{figure}
    \centering
    \includegraphics[width=1.0\linewidth]{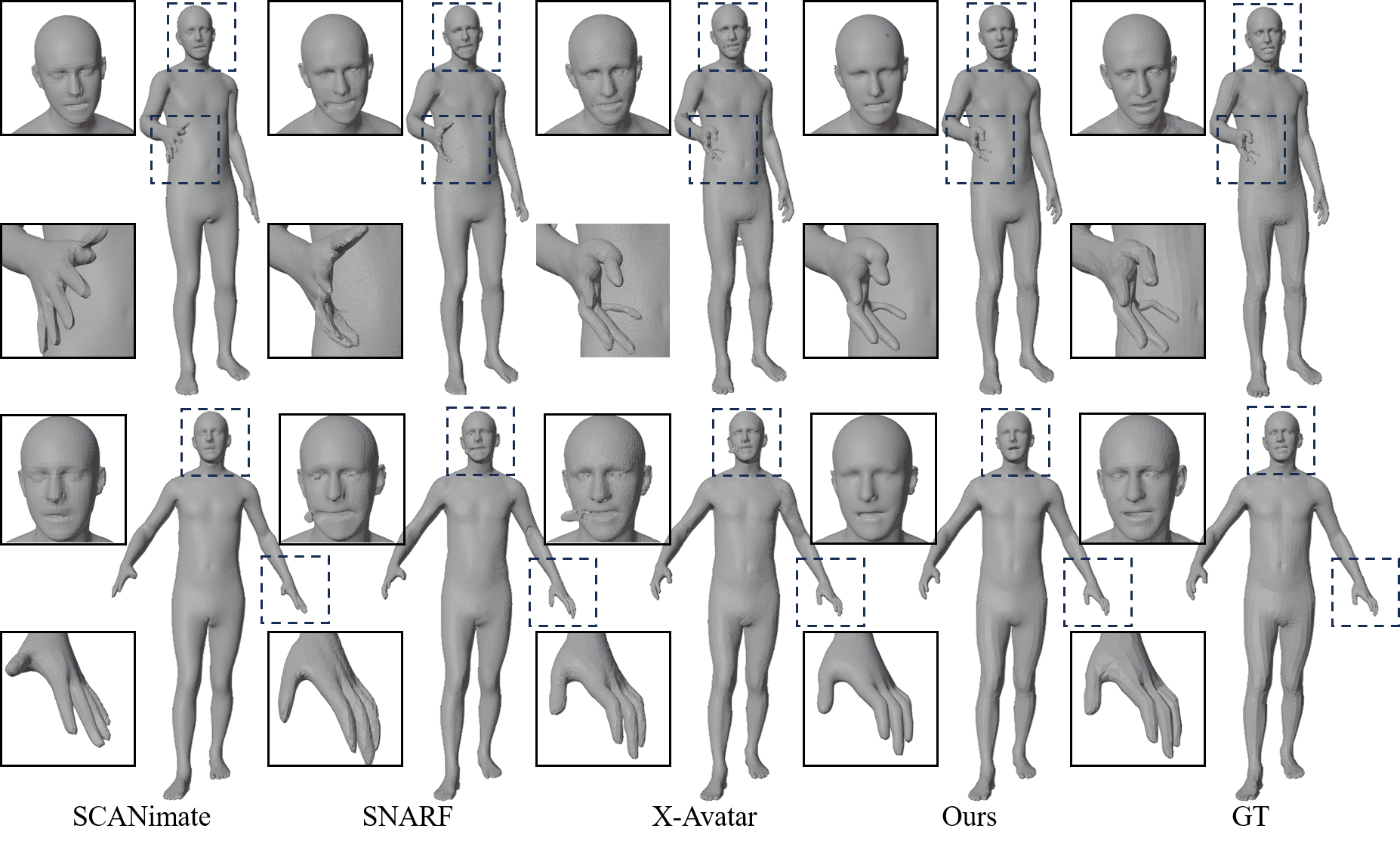}
\caption{\textbf{Qualitative Comparison on GRAB dataset} Our method outperforms the implicit methods~\cite{DBLP:conf/cvpr/scanimate21,DBLP:conf/iccv/snarf21,shen2023xavatar}.}
    \label{fig:grab}
\end{figure}

\subsection{Results on GRAB dataset}

GRAB dataset~\cite{DBLP:conf/eccv/grab20} consists of minimally clothed SMPL-X meshes with diverse hand poses and facial expressions. We follow the same partition strategy as X-Avatar for evaluation. We evaluate the geometric accuracy via volumetric IoU, Chamfer distance~(CD) and normal consistency~(NC) metrics. Volumetric IoU provides a measure of the overlap between the reconstructed mesh and the ground-truth mesh. Chamfer distance quantifies the reconstruction accuracy comparing to the ground-truth mesh. Normal consistency evaluates the fineness of reconstructed local details. We compare our proposed method against recent implicit human avatar methods including SCANimate~\cite{DBLP:conf/cvpr/scanimate21}, SNARF~\cite{DBLP:conf/iccv/snarf21} and X-avatar~\cite{shen2023xavatar}.

Table~\ref{tab:grab} presents the quantitative results on GRAB dataset. Our method demonstrates comparable performance with X-Avatar and surpasses SCANimate and SNARF. SCANimate and SNARF exhibit subpar performance when it comes to face and hand reconstruction. Fig.~\ref{fig:grab} shows the qualitatively results. Both Xavatar and our method excel in accurately reconstructing both facial expressions and hand gestures. SCANimate recovers a mean hand and SNARF cannot handle complex hand poses, resulting in the large errors.

\begin{table*}
	\centering
	\caption{Quantitative results on X-avatar dataset.}
	\begin{tabular}{lcc|cc|cc|cc|cc}
		\toprule
  \multirow{2}{*}{Method} & \multicolumn{2}{c}{CD$\downarrow$}  & \multicolumn{2}{c}{CD-MAX $\downarrow$}  & \multicolumn{2}{c}{NC $\uparrow$} & \multicolumn{2}{c}{IoU $\uparrow$ } & \multicolumn{2}{c}{Time $\downarrow$} \\
  \cline{2-11} & All & Hands & All & Hands & All & Hands & All & Hands & Training & Inference\\
		\midrule
        SCANimate~\cite{DBLP:conf/cvpr/scanimate21} & 6.54 & 9.78 & 59.71 & 48.32 & 0.925 & 0.726 & 0.919 & 0.55 & 30 h & 3.0 s \\
		SNARF~\cite{DBLP:conf/iccv/snarf21} & 5.05 & 7.23 & 55.06 & 37.15 & 0.934 & 0.788 & 0.937 & 0.608 & 18 h & 3.5 s \\
		X-avatar~\cite{shen2023xavatar} & 4.43 & 5.14 & 47.56 & 22.15 & 0.939 & 0.793 & 0.965 & 0.776 & 25 h & 4.1 s \\
		Ours & \textbf{4.30} & \textbf{4.96} & \textbf{44.36} & \textbf{20.22} & \textbf{0.945} & \textbf{0.832} & \textbf{0.968} & \textbf{0.787} & \textbf{10} h & \textbf{0.1} s \\
		\bottomrule
	\end{tabular}
	\label{tab:xhumans}
\end{table*}

\begin{figure*}
    \centering
    \includegraphics[width=1.0\linewidth]{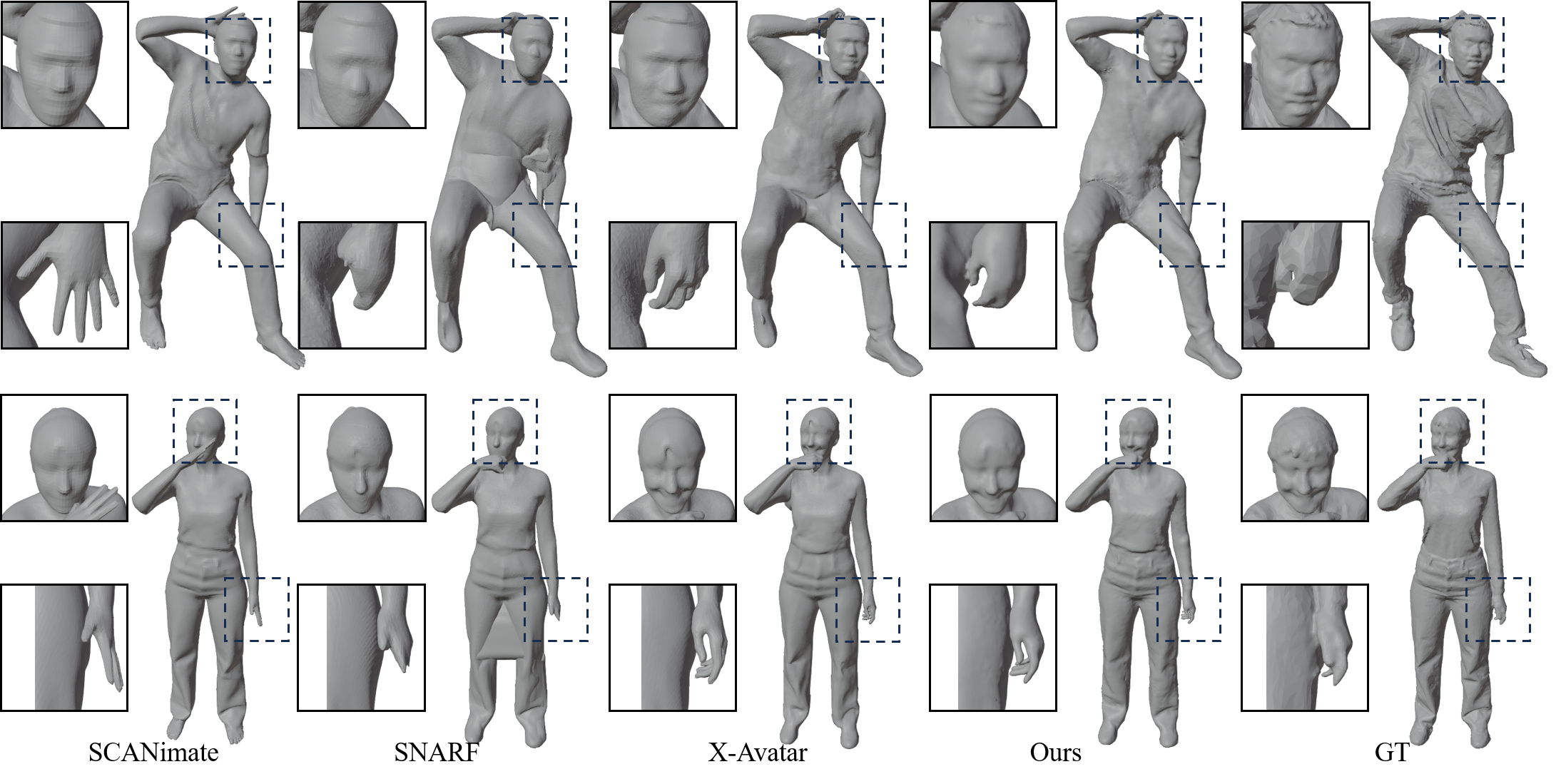}
    \caption{\textbf{Qualitative Comparison on X-Humans dataset} Our method outperforms these implicit methods.}
    \label{fig:xhumans}
\end{figure*}

\subsection{Results on X-Humans}

X-Humans~\cite{shen2023xavatar} dataset comprises textured 3D clothed scans of humans, encompassing a wide range of body poses, hand gestures and facial expressions. These scans are captured using a multi-view volumetric capture stage. A custom registration pipeline effectively extracts the SMPL-X parameters for each scan. We adopt the identical  evaluation metrics used in GRAB dataset. We also conduct a comparison of our approach against SCANimate, SNARF and X-avatar.

The quantitative results obtained from the X-Humans dataset are summarized in Table~\ref{tab:xhumans}. Our proposed approach  successfully attains the lowest Chamfer distance, highest normal consistency and volumetric IoU, demonstrating exceptional performance in all three metrics. Fig.~\ref{fig:xhumans} shows the qualitatively results. Similar to the results on GRAB dataset, both SCANimate and SNARF fall short in accurately capturing hand poses and facial expressions, resulting in erroneous face and hand results. Although X-Avatar is able to recover correct facial expressions and hand poses, implicit representations may lead to artifacts in some extreme poses. Furthermore, it has limited ability to recover clothing details.

\noindent\textbf{Efficiency Comparison} Our proposed method has advantages in terms of both training time and inference time. SCANimate acquires the consistency between canonical space and pose space conversions in a weakly supervised manner, which requires additional optimization of an inverse LBS network. To enable weak supervision, SCANimate needs to pre-train the LBS network and perform linear blend skinning twice during training. While it takes long time for SNARF and X-Avatar to find canonical correspondences which satisfy the forward skinning equation via iterative root finding. To enable part-aware initialization, X-Avatar dedicates extra time during the training phase to compute the category of each point. During inference, these implicit methods need to query all grid positions to obtain occupancy value and employ marching cube~\cite{DBLP:conf/siggraph/Marchingcubes87} to extract the final mesh. While multi-scale methods can help alleviate the computational load, there remains a considerable number of SDF values that must be calculated at various locations. By taking advantage of oriented points representation, we can obtain the final mesh efficiently through Poisson reconstruction. It takes 0.8 seconds when we adopt traditional CPU-based Poisson reconstruction. By employing the GPU-based Poisson solver~\cite{Peng2021SAP} and torchmucbes\footnote{https://github.com/tatsy/torchmcubes}, the processing time decreases to 0.1 seconds. All the experiments about computational time are conducted on the same machine with a single NVIDIA 3090Ti GPU. The dimension of marching cube used by implicit methods and the depth of Poisson reconstruction are $512$ and $9$ respectively.

\begin{table}
    \centering
    \caption{Quantitative Ablation Study on the Proposed Loss}
    \resizebox{1.00\linewidth}{!}{
    \begin{tabular}{lcc|cc|cc|cc}
    \toprule
    \multirow{2}{*}{Method} & \multicolumn{2}{c}{CD$\downarrow$}  & \multicolumn{2}{c}{CD-MAX $\downarrow$}  & \multicolumn{2}{c}{NC $\uparrow$} & \multicolumn{2}{c}{IoU $\uparrow$ } \\
    \cline{2-9} & All & Hands & All & Hands & All & Hands & All & Hands\\
    \midrule
    w/o emd & 4.38 & 5.01 & 45.93 & 20.64 & 0.940 & 0.818 & 0.962 & 0.779 \\
    w/o chamfer & 4.36 & 5.00 & 44.50 & 20.28 & 0.943 & 0.820 & 0.964 & 0.782 \\
    w/o normal & 4.39 & 5.18 & 45.83 & 21.24 & 0.935 & 0.805 & 0.965 & 0.784 \\
    w/o reg & 4.43 & 5.27 & 45.53 & 23.78 & 0.940 & 0.823 & 0.960 & 0.770 \\
    \bottomrule
    \end{tabular}}
    \label{tab:abl}
\end{table}

\subsection{Ablation Studies}

We have performed comprehensive ablation studies to analyze the impact of our proposed loss functions. The quantitative results are presented in Table~\ref{tab:abl}. The EMD loss mitigates issues related to uneven point density distribution and preserves fine details, thereby enhancing the overall quality of the reconstruction. The uniform distribution facilitated by the EMD loss also enhance better appearance recovery because we utilize KNN to interpolate the neural textures. The Chamfer distance loss effectively minimizes discrepancies between the sampled points and posed points, resulting in improved alignment. Additionally, the normal loss contributes to enhancing normal consistency and facilitating the recovery of intricate clothing details. The LBS regularization loss aids in the precise recovery of hand shapes and reduces the Chamfer distance within the hand region.

\begin{figure*}
    \centering
    \begin{tabular}{@{\hskip2pt}c@{\hskip2pt}@{\hskip2pt}c@{\hskip2pt}@{\hskip2pt}c@{\hskip2pt}@{\hskip2pt}c@{\hskip2pt}@{\hskip2pt}c@{\hskip2pt}@{\hskip2pt}c@{\hskip2pt}@{\hskip2pt}c@{\hskip2pt}c@{\hskip2pt}c@{\hskip2pt}}
    \includegraphics[width=0.12\textwidth,trim=250 0 250 0,clip]{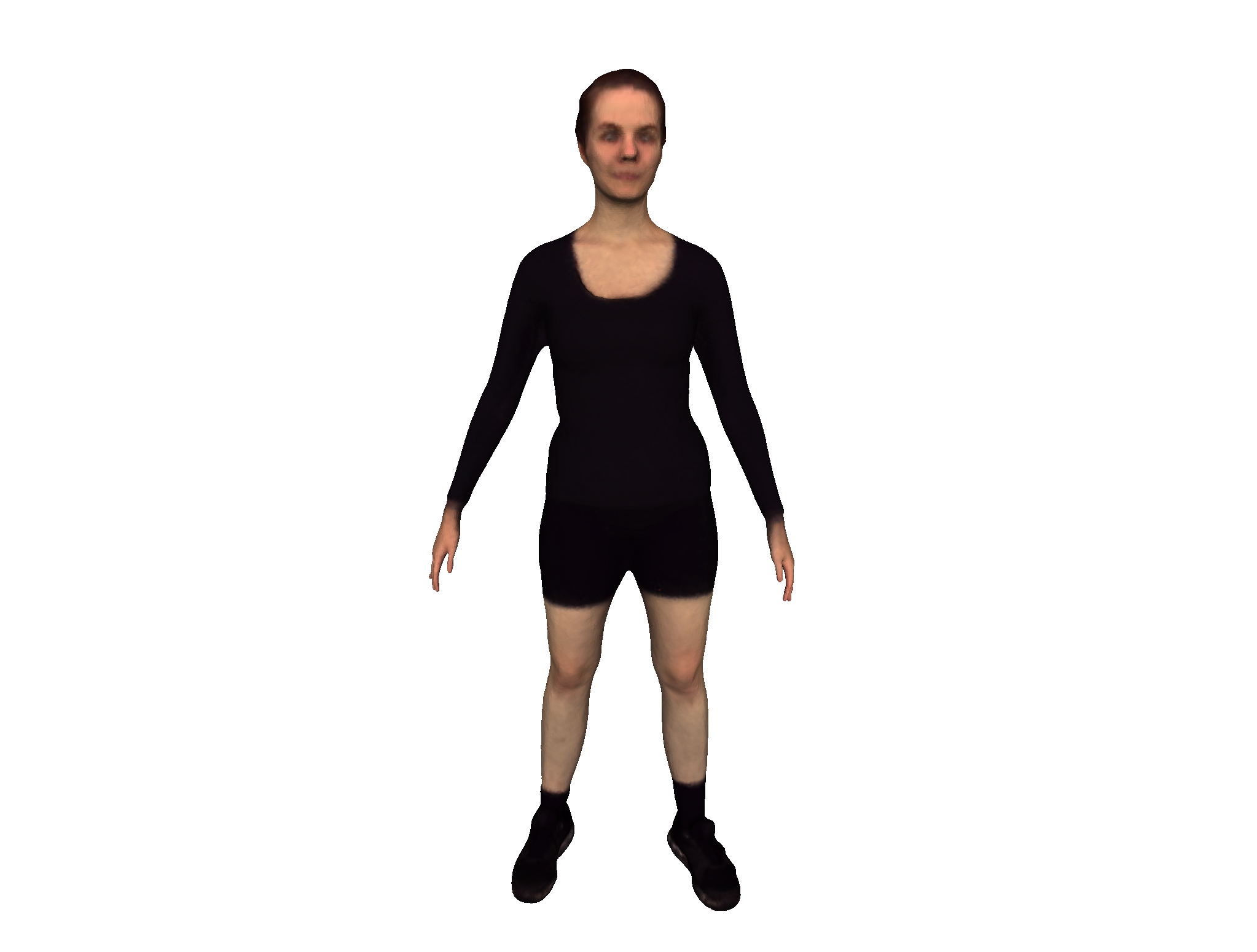} & 
    \includegraphics[width=0.12\textwidth,trim=275 0 225 0,clip]{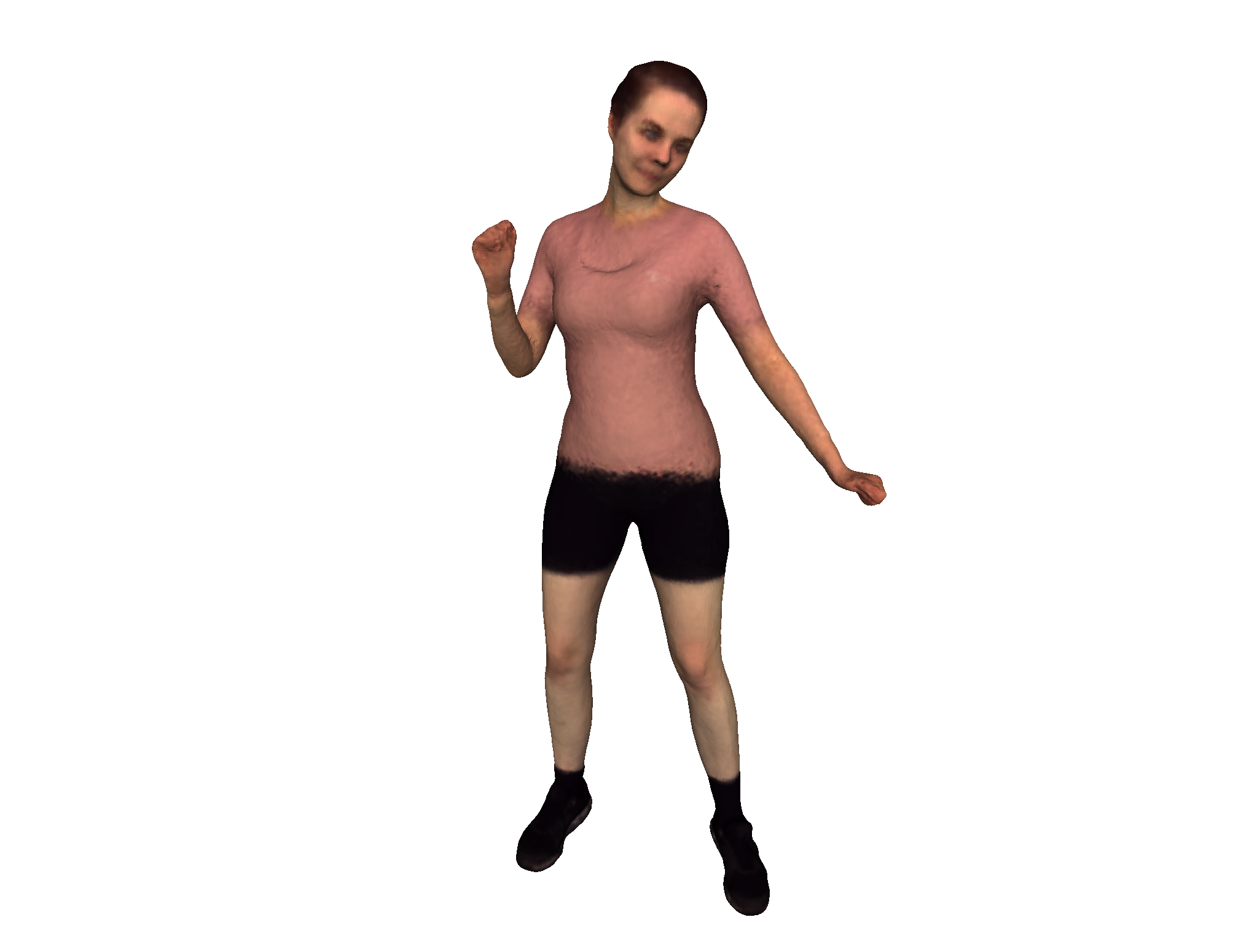} &
    \includegraphics[width=0.12\textwidth,trim=250 0 250 0,clip]{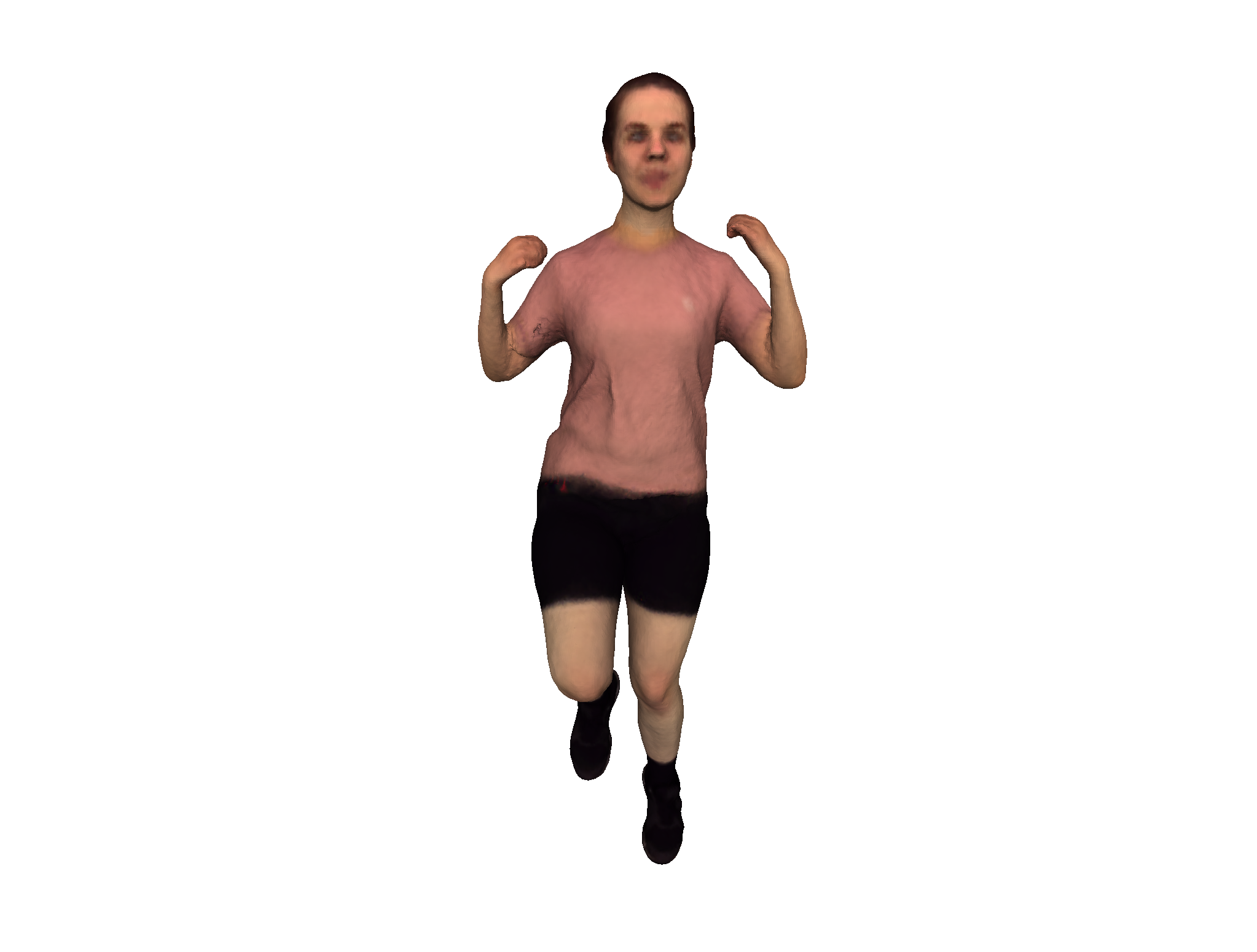} &
    \includegraphics[width=0.12\textwidth,trim=250 2 250 0,clip]{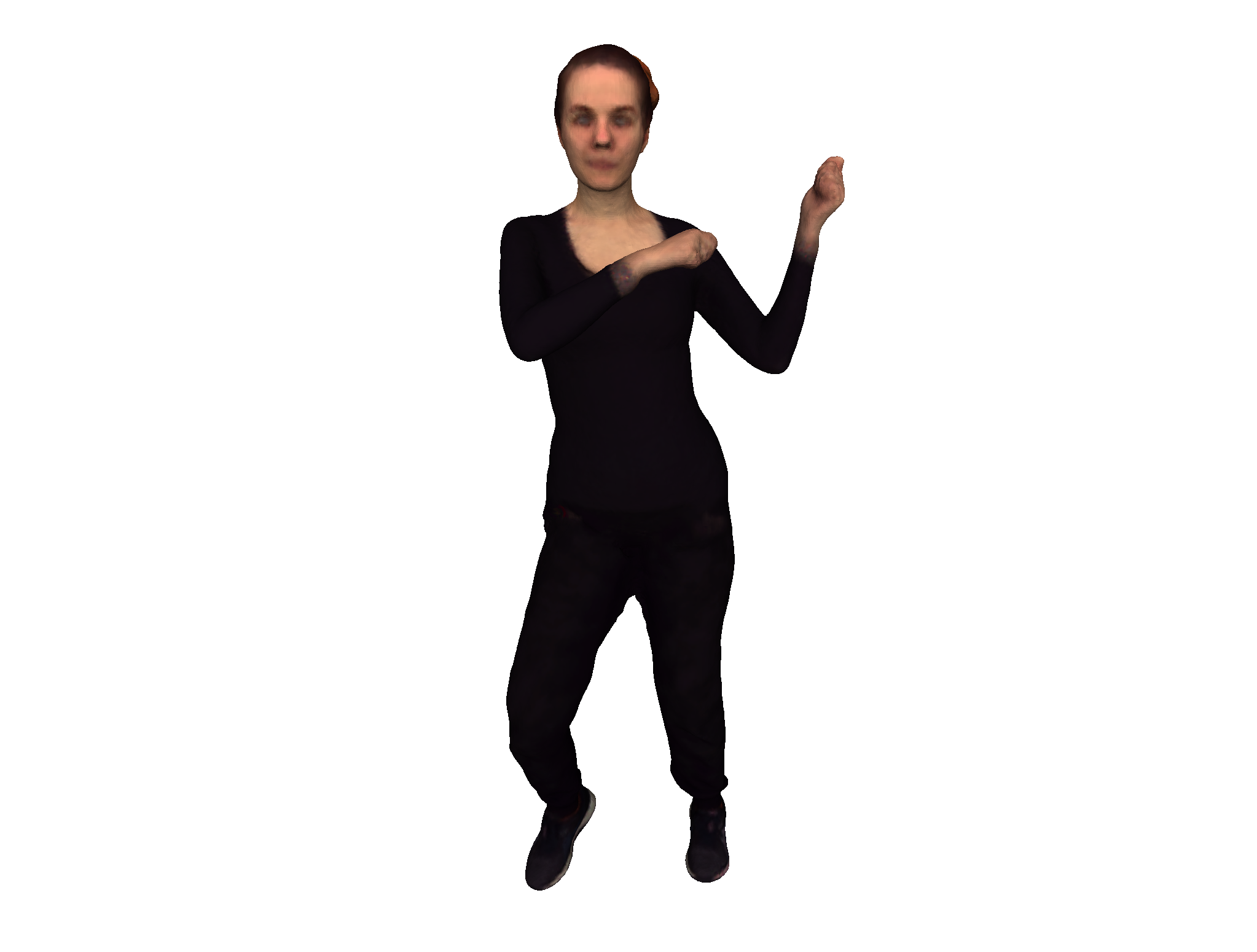} &
    \includegraphics[width=0.12\textwidth,trim=250 0 250 0,clip]{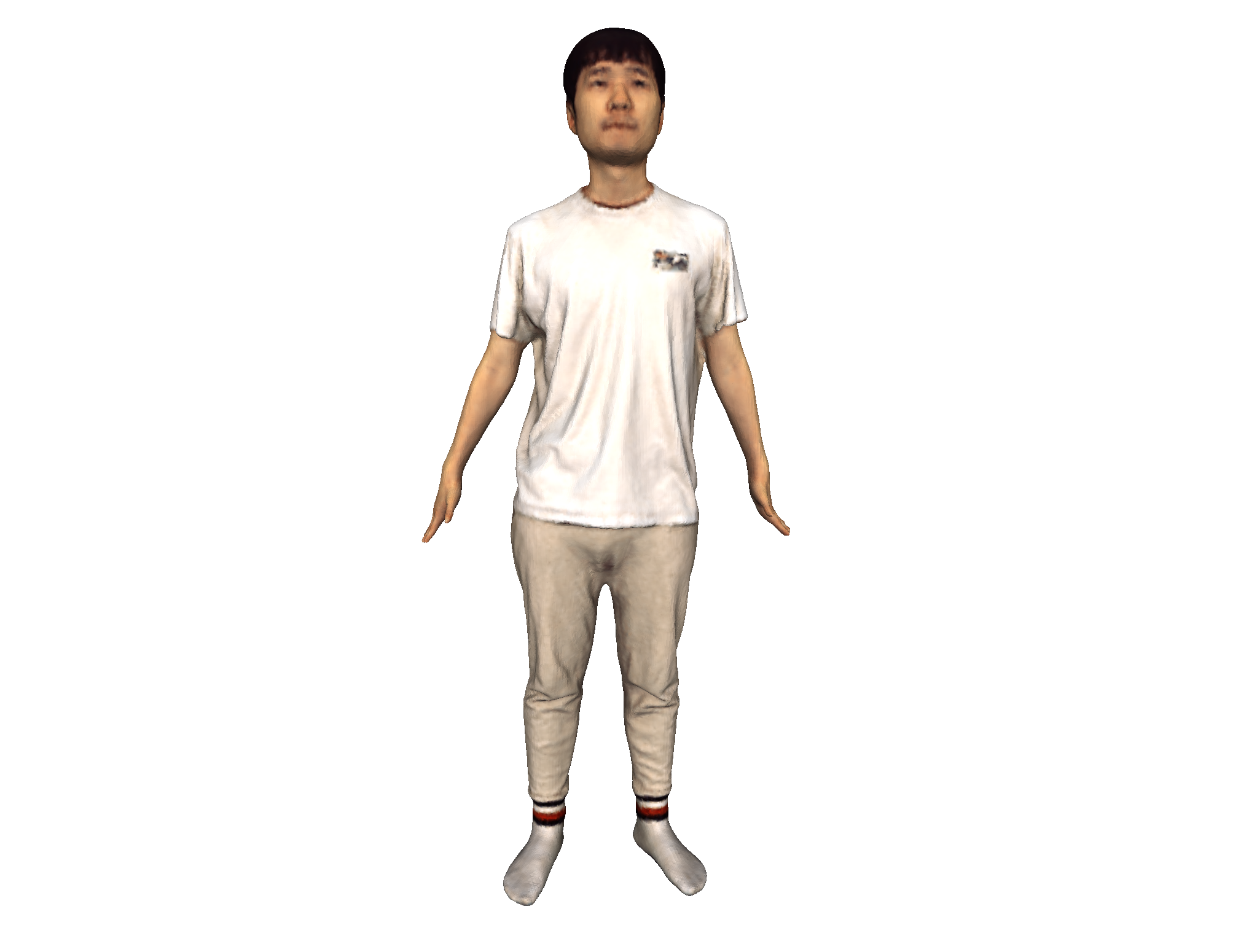} &
    \includegraphics[width=0.12\textwidth,trim=250 0 250 0,clip]{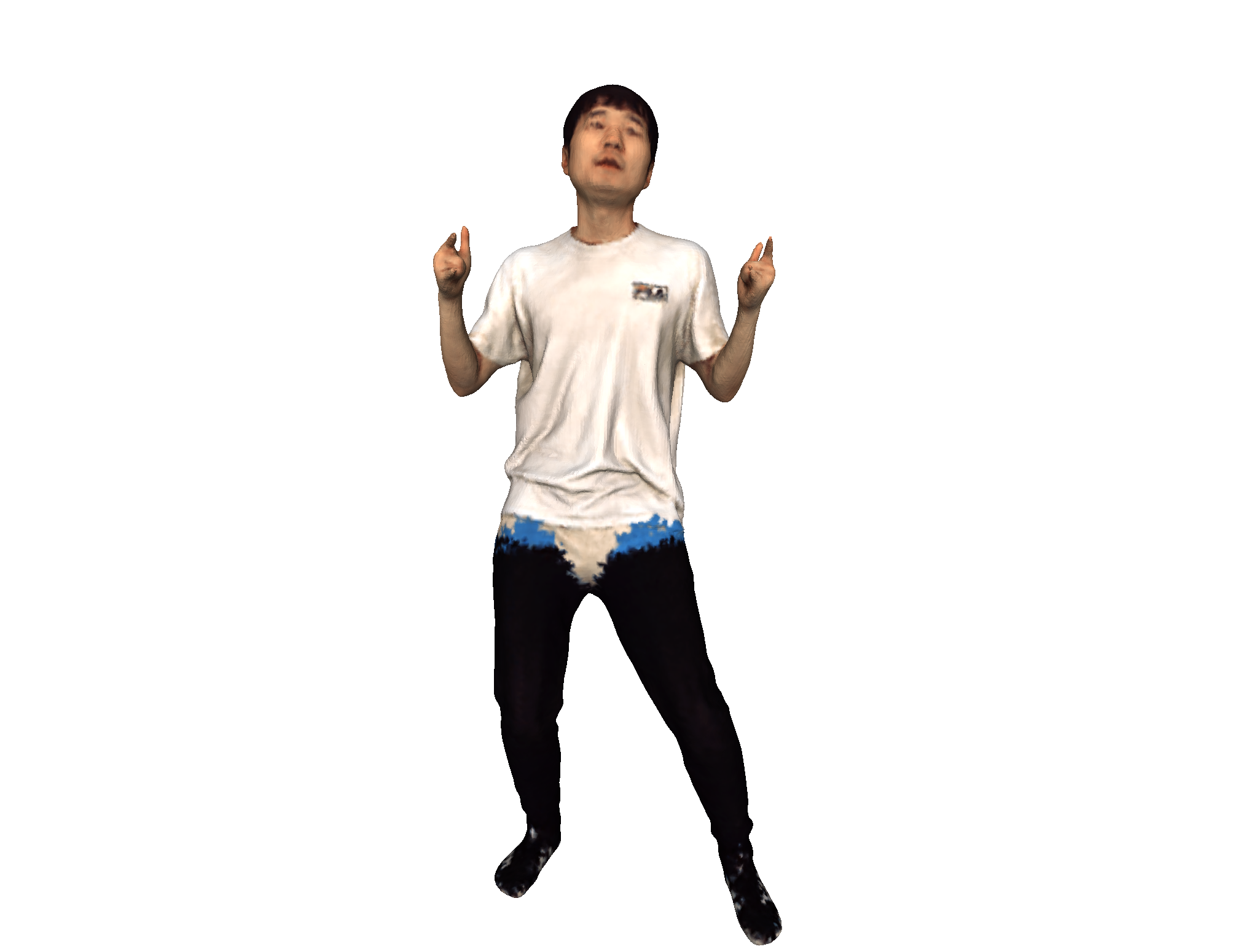} &
    \includegraphics[width=0.12\textwidth,trim=250 0 250 0,clip]{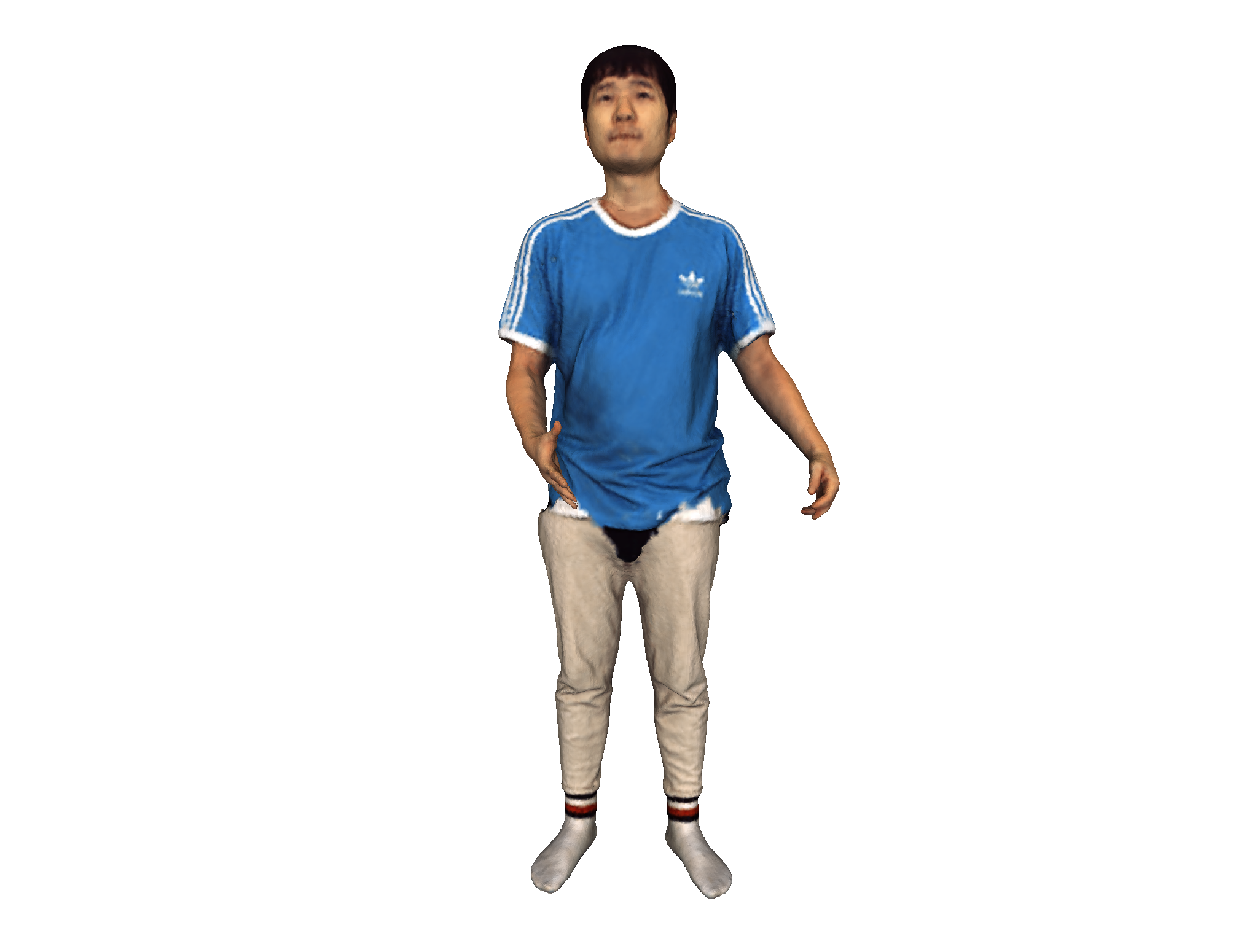} &
    \includegraphics[width=0.12\textwidth,trim=250 0 250 0,clip]{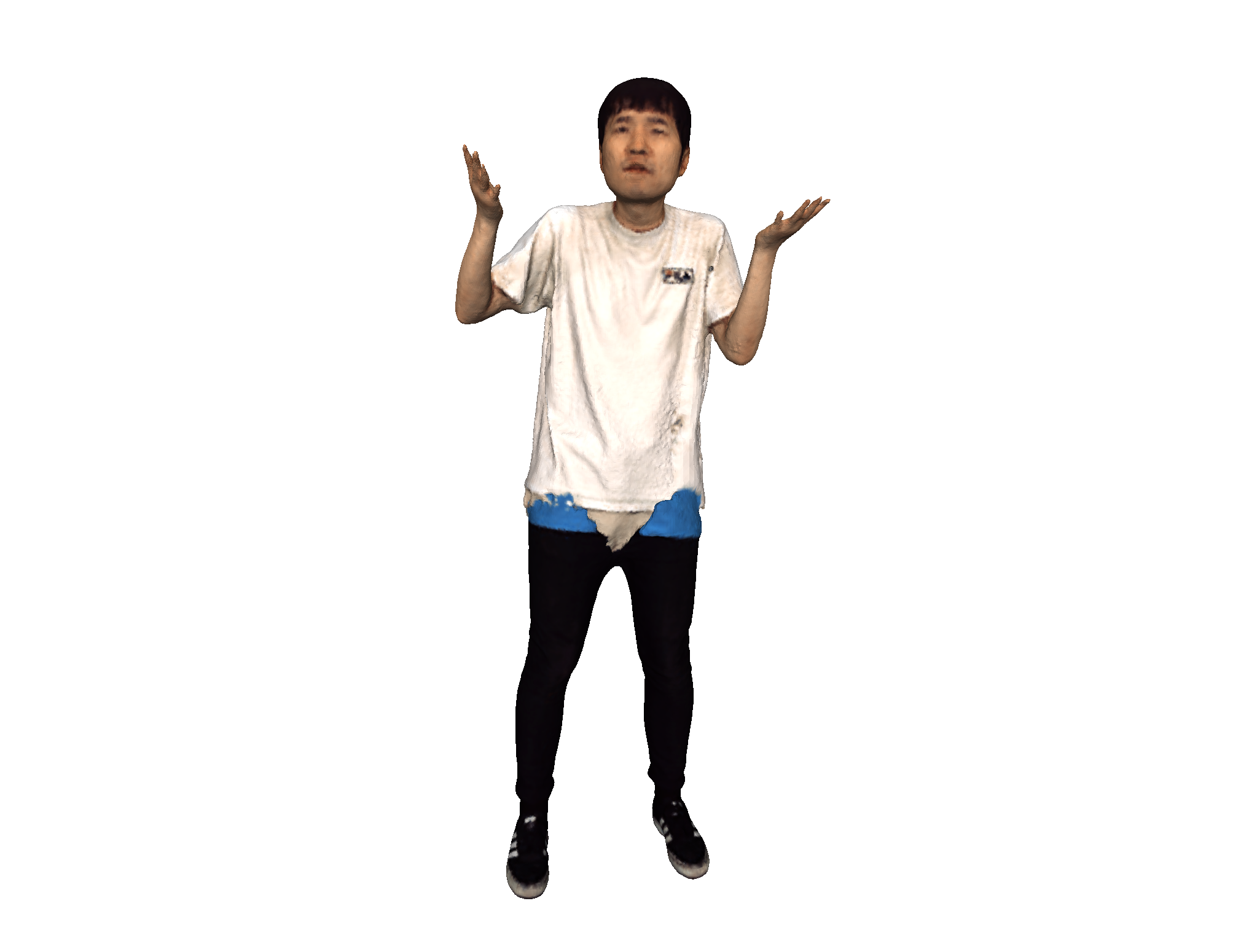} \\ 
    \includegraphics[width=0.12\textwidth,trim=250 0 250 0,clip]{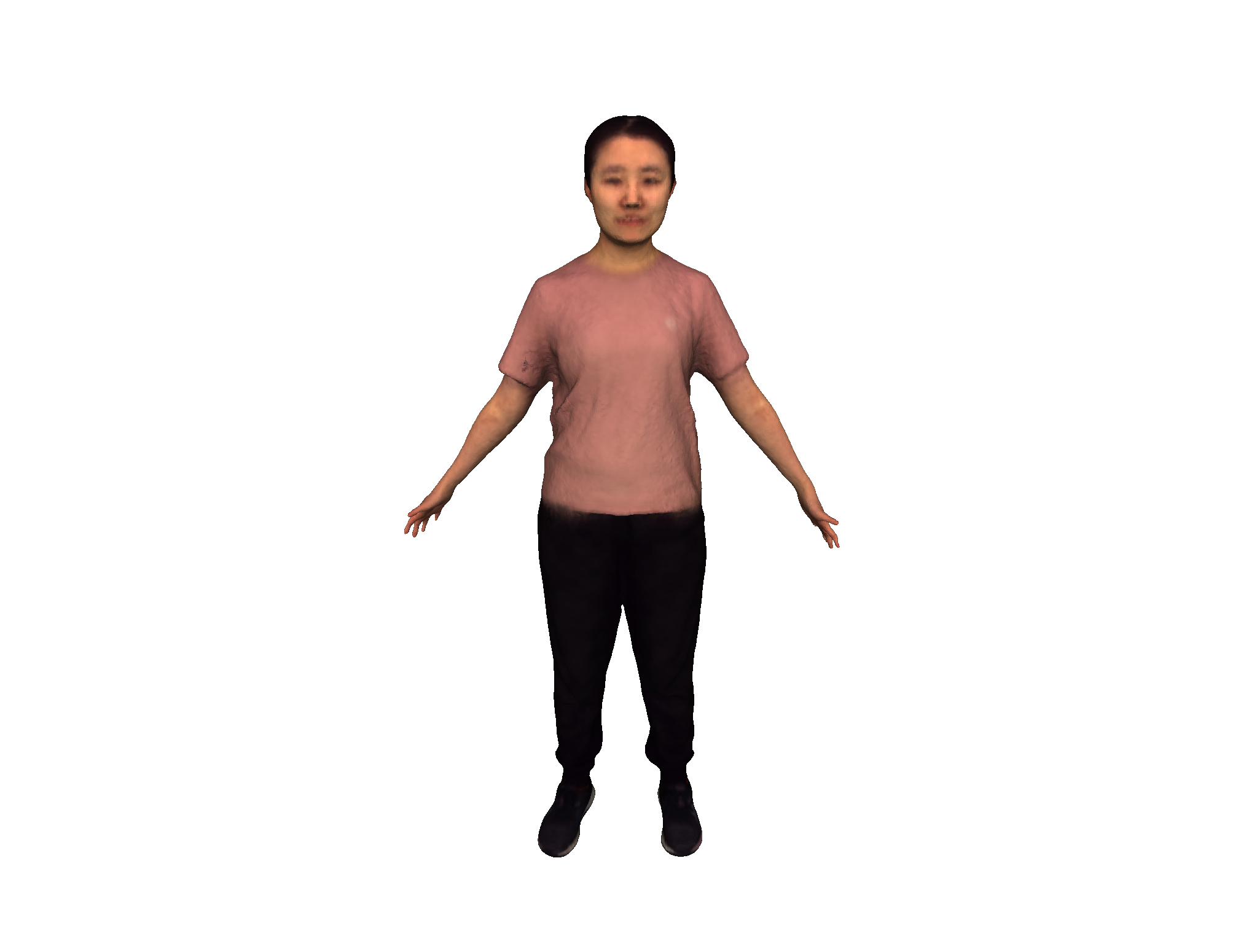} &
    \includegraphics[width=0.12\textwidth,trim=250 0 250 0,clip]{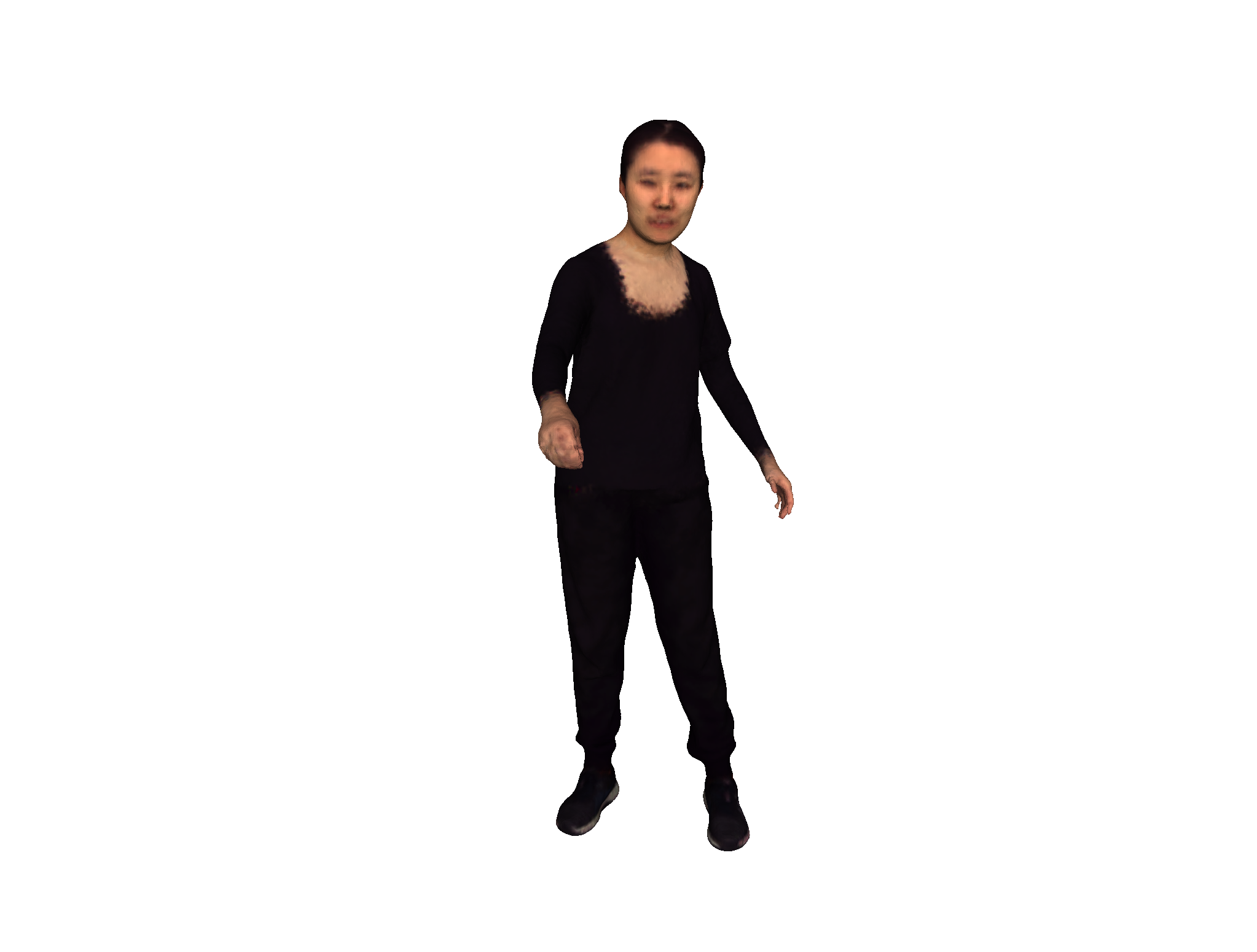} &
    \includegraphics[width=0.12\textwidth,trim=250 0 250 0,clip]{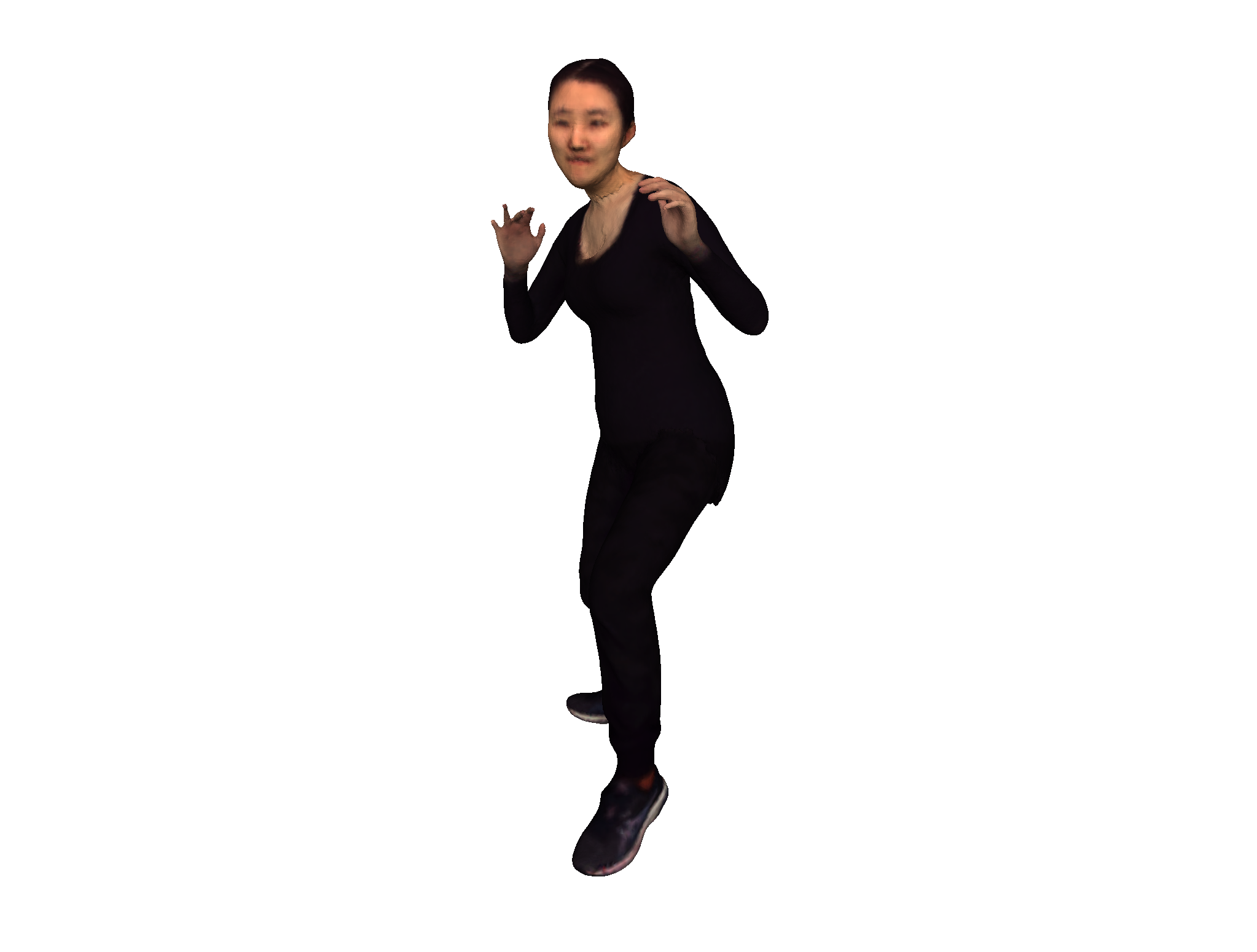} &
    \includegraphics[width=0.12\textwidth,trim=290 0 210 0,clip]{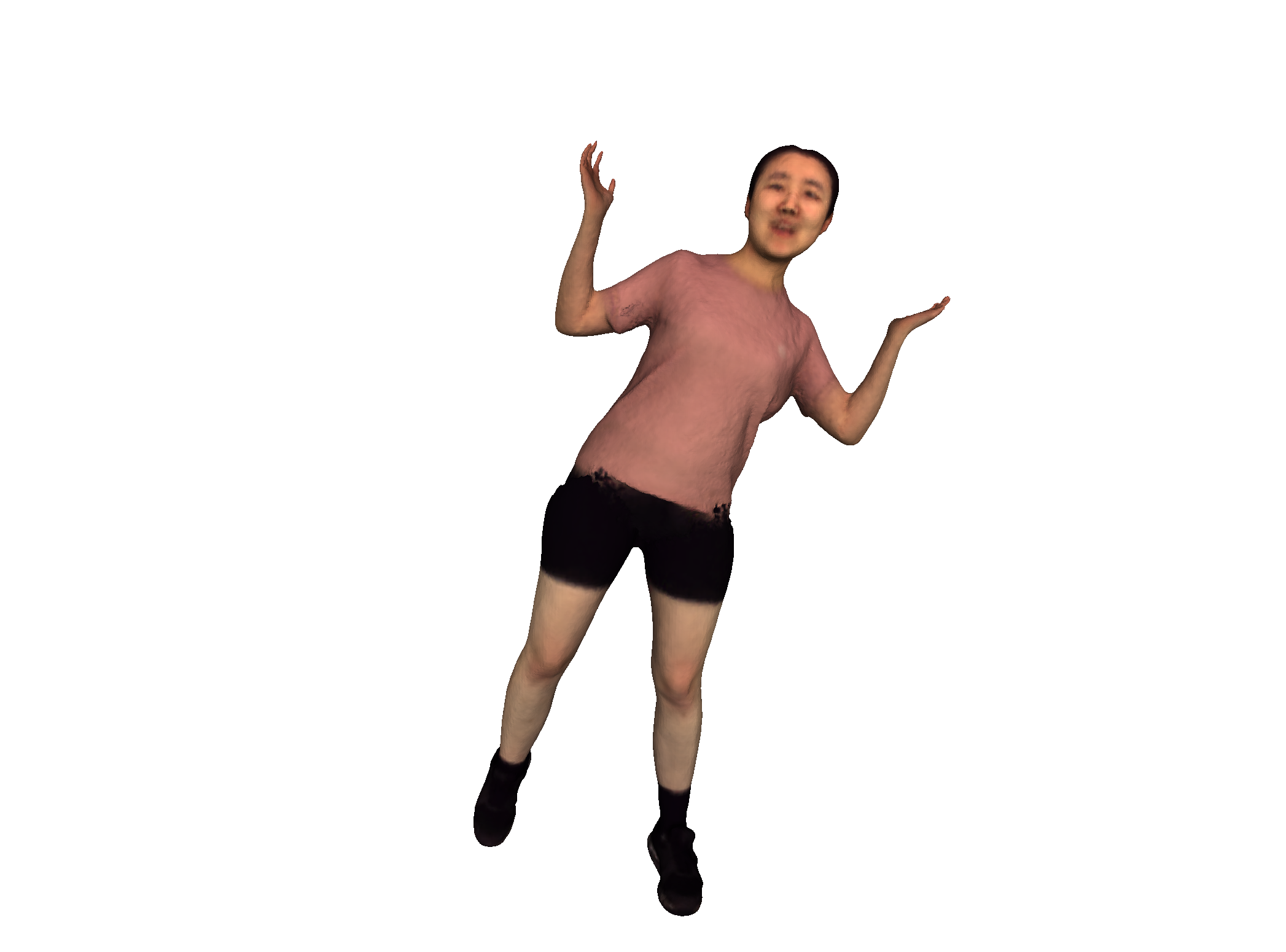} &
    \includegraphics[width=0.12\textwidth,trim=250 0 250 0,clip]{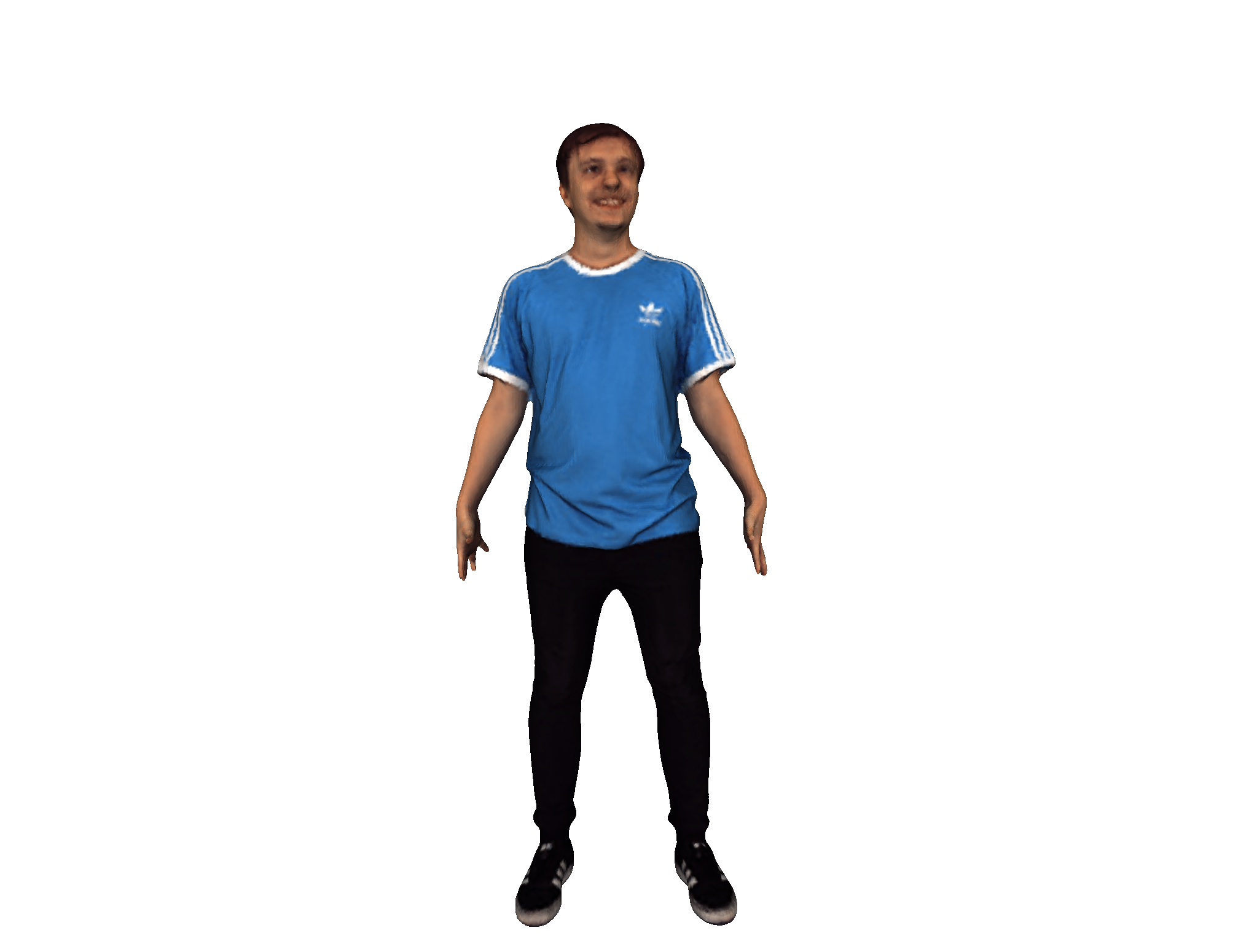} &
    \includegraphics[width=0.12\textwidth,trim=250 0 250 0,clip]{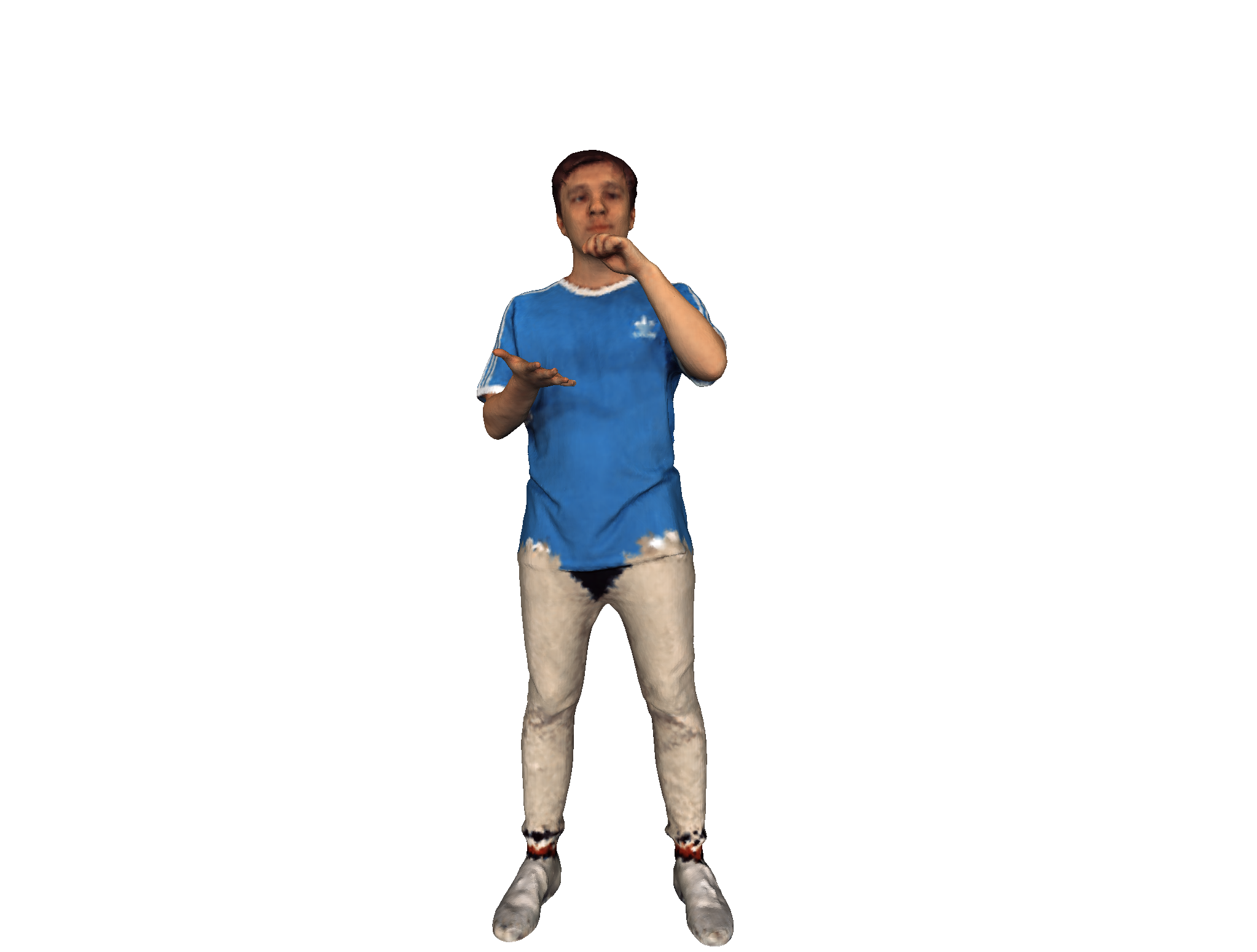} &
    \includegraphics[width=0.12\textwidth,trim=250 0 250 0,clip]{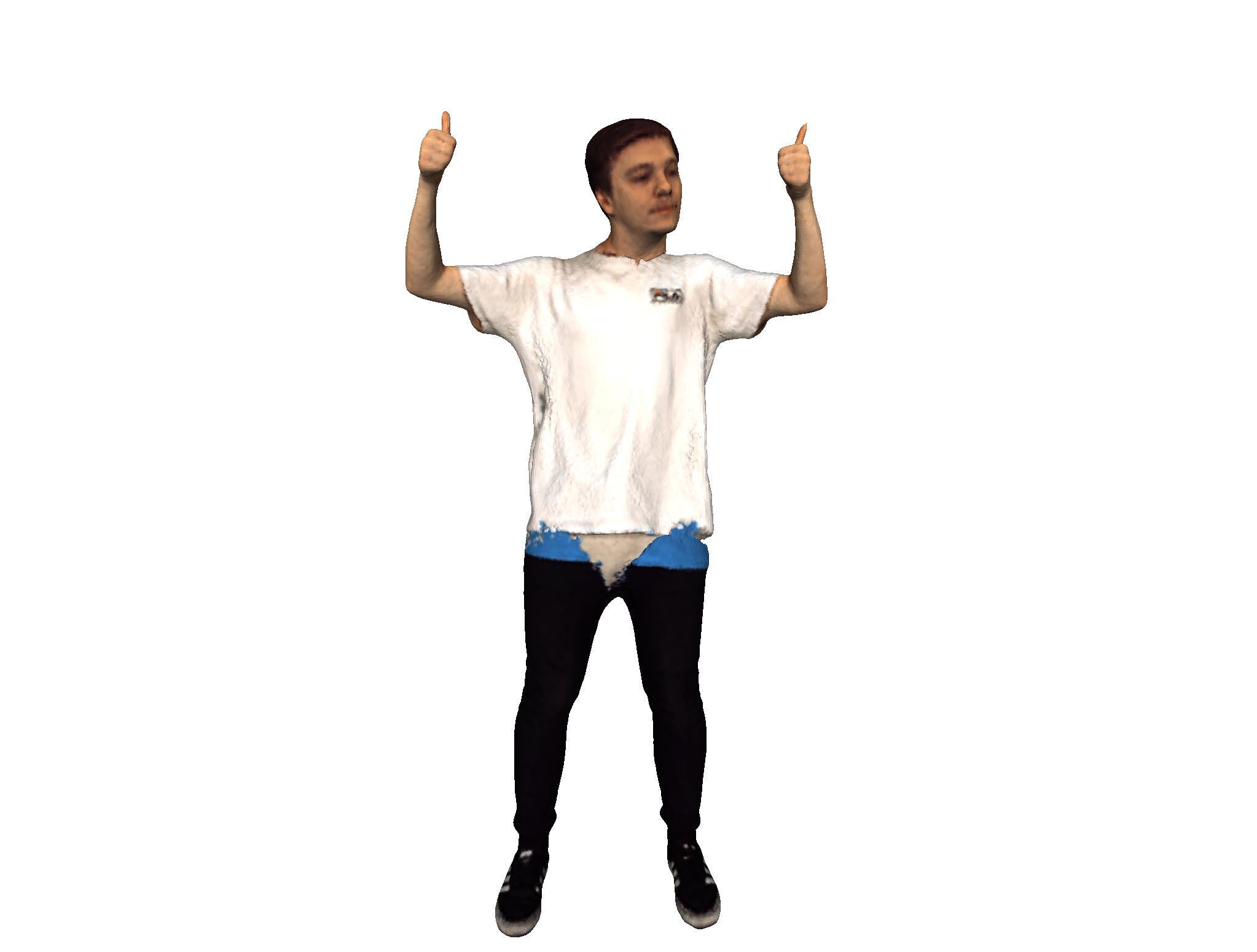} &
    \includegraphics[width=0.12\textwidth,trim=250 0 250 0,clip]{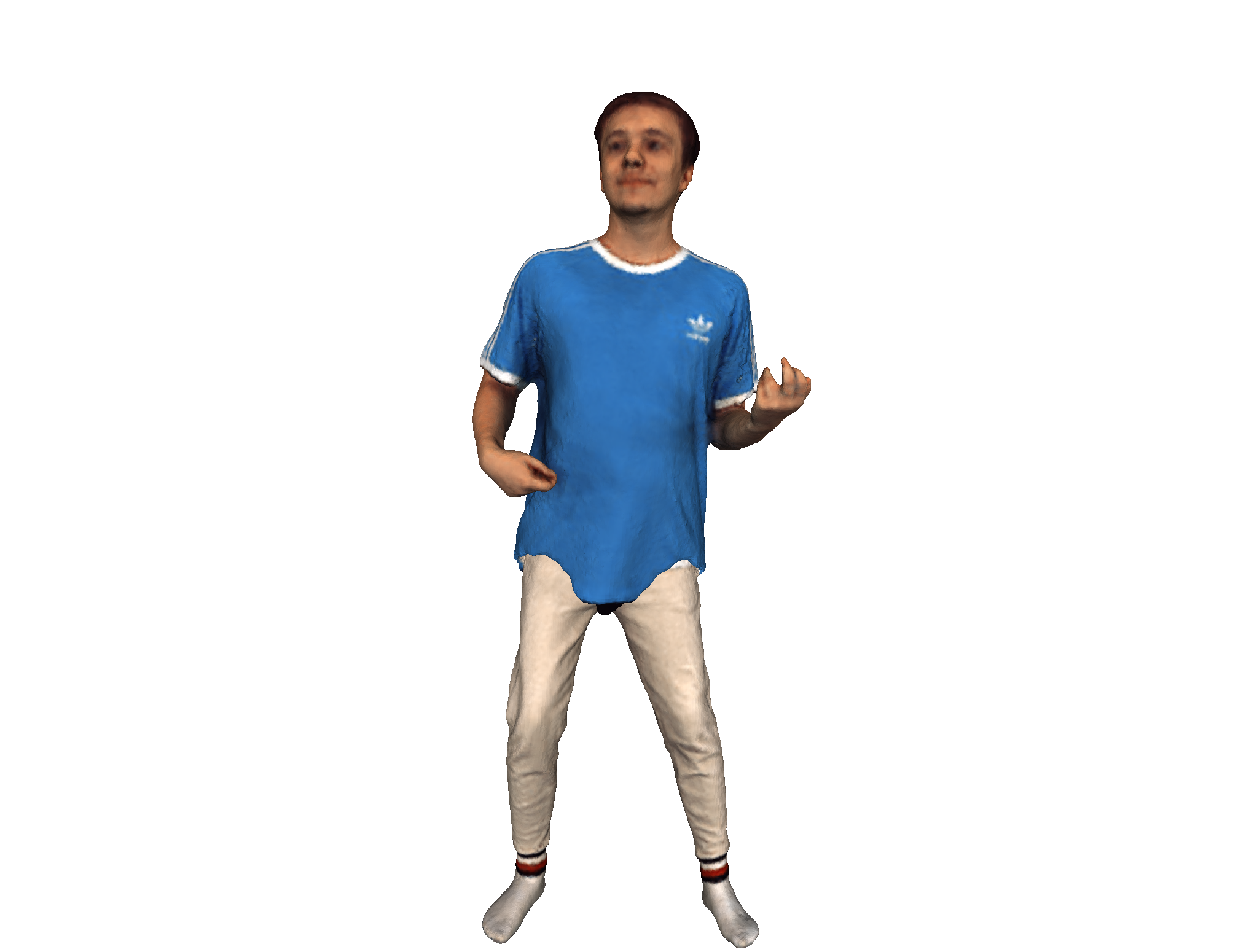} \\
    \small{Ori Mesh} & \small{Texture transfer} & \small{Points transfer} & \small{Points transfer} & \small{Ori Mesh} & \small{Texture transfer} & \small{Points transfer} & \small{Points transfer}
    \end{tabular}
    \caption{\textbf{Composition results of different avatars on X-Humans Dataset.} We show qualitative results of composited avatars. We can transfer only the texture or both the texture and the point clouds.}
    \label{fig:composite}
\end{figure*}

\subsection{Avatar Composition}

Another advantage of oriented points representation is its ability to preserve semantic information for avatar composition. Despite utilizing part-aware initialization and sampling strategies, implicit methods fall short in obtaining the semantic information of the reconstructed human meshes during inference. Every vertex is handled without discrimination. As a consequence, implicit methods need to train a distinct model for each subject.

Our proposed novel method enriches points with semantic information from SMPL-X model, enabling us to identify the category associated with each vertex in the reconstructed meshes. Utilizing these categorizations, we can conduct a more thorough analysis of human movements. The inherent benefits of Poisson reconstruction enable us to seamlessly merge disparate points. Through exchanging textures and points belonging to the same category across different individuals, our approach enables virtual try-on and the composition of human avatars. To preserve the clothing geometry, we can either exclusively interchange the neural texture of the points or simultaneously swap both the points and texture to create a composite human avatar. Fig.~\ref{fig:composite} shows the results of composited avatars. The promising results indicate that our method is capable of generating realistic human bodies in both texture transfer and points transfer scenarios.

\section{Limitations and Conclusions}

By transferring the semantic information of the SMPL-X model to the clothed avatar,  our proposed method encounters challenges in handling loose clothing, such as long skirts. Moreover, Poisson reconstruction~\cite{DBLP:conf/sgp/psr06} aims to create a surface that minimizes the differences in normals at neighboring vertices, which tends to produce smoothing meshes.

In this paper, we introduce an efficient point based human avatar that comprehensively captures poses, expressions and appearances. We utilize two MLPs to model pose-dependent deformation and estimate LBS weights. Poisson reconstruction provides an efficient way to transform the oriented points into meshes in a timely manner. Comparing to implicit representation, our oriented point clouds representation provides interpretability to the avatar model. Furthermore, we propose a novel method for transferring the semantic information from SMPL-X model to the point clouds, which enables to create avatars by compositing different subjects. Experimental results demonstrate that our proposed approach performs comparable to the state-of-the-art implicit approaches while requiring less training and inference time.

{\small
\bibliographystyle{ieee_fullname}
\bibliography{egbib}
}

\end{document}